\documentclass[10pt,journal,compsoc]{IEEEtran}

\ifCLASSOPTIONcompsoc
  \usepackage[nocompress]{cite}
  \usepackage{multirow}
  \usepackage{amssymb}
  \usepackage{graphicx}
  \usepackage{amsmath}
  \usepackage{multirow}
  \usepackage{multicol}
  \usepackage{array}
  \usepackage{makecell}
  \usepackage{colortbl}
  
  \usepackage{amssymb}
  \usepackage{booktabs}
  \usepackage{multirow}
  \usepackage{bm}
  \usepackage{bbding}

  \usepackage[colorlinks, linkcolor=red]{hyperref}
  
  \usepackage{wrapfig} 
  \usepackage{xcolor} 
  \usepackage{soul}
\else
  \usepackage{cite}
\fi

\usepackage{tablefootnote}
\usepackage{footnotehyper}

\newcommand{\ourMethod}{UniLION}

\begin{document}
%

\title{{\ourMethod}: Towards Unified Autonomous Driving Model with Linear Group RNNs}

\author{Zhe Liu$^{1,3}$, Jinghua Hou$^{1}$, Xiaoqing Ye$^{4}$, Jingdong Wang$^{4}$, \textit{Fellow, IEEE}, Hengshuang Zhao$^{3,\dagger}$, \textit{Member, IEEE}, Xiang Bai$^{2,\dagger}$, \textit{Senior Member, IEEE}
\thanks{$^{1}$School of Electronic Information and Communications, Huazhong University of Science and Technology~(HUST), Wuhan, China.}
\thanks{$^{2}$School of Software
Engineering, Huazhong University of Science and Technology~(HUST), Wuhan, China.}
\thanks{$^{3}$Department of Computer Science, The University of Hong Kong, Hong Kong~(HKU) SAR, China.}
\thanks{$^{4}$Baidu Inc., Beijing, China.}
\thanks{$^{\dagger}$Corresponding authors: Hengshuang Zhao (e-mail: hszhao@cs.hku.hk) and Xiang Bai (e-mail: xbai@hust.edu.cn).}
}


\IEEEtitleabstractindextext{%
\begin{abstract}
Although transformers have demonstrated remarkable capabilities across various domains, their quadratic attention mechanisms introduce significant computational overhead when processing long-sequence data. In this paper, we present a unified autonomous driving model, {\ourMethod}, which efficiently handles large-scale LiDAR point clouds, high-resolution multi-view images, and even temporal sequences based on the linear group RNN operator~(\textit{i.e.}, performs linear RNN for grouped features). Remarkably, {\ourMethod} serves as a single versatile architecture that can seamlessly support multiple specialized variants~(\textit{i.e.}, LiDAR-only, temporal LiDAR, multi-modal, and multi-modal temporal fusion configurations) without requiring explicit temporal or multi-modal fusion modules.
Moreover, {\ourMethod} consistently delivers competitive and even state-of-the-art performance across a wide range of core tasks, including 3D perception (\textit{e.g.}, 3D object detection, 3D object tracking, 3D occupancy prediction, BEV map segmentation), prediction (\textit{e.g.}, motion prediction), and planning (\textit{e.g.}, end-to-end planning). This unified paradigm naturally simplifies the design of multi-modal and multi-task autonomous driving systems while maintaining superior performance. Ultimately, we hope {\ourMethod} offers a fresh perspective on the development of 3D foundation models in autonomous driving. Code is available at {\textcolor{magenta}{\url{https://github.com/happinesslz/UniLION}}}
\end{abstract}

\begin{IEEEkeywords}
Unified Model, Linear Group RNNs, Autonomous Driving, 3D Perception, Motion Prediction, Planning
\end{IEEEkeywords}}

\maketitle

\IEEEdisplaynontitleabstractindextext

%
\IEEEpeerreviewmaketitle

\section{Introduction}

\begin{figure}[h!]
\centering
\includegraphics[width=0.99\linewidth]{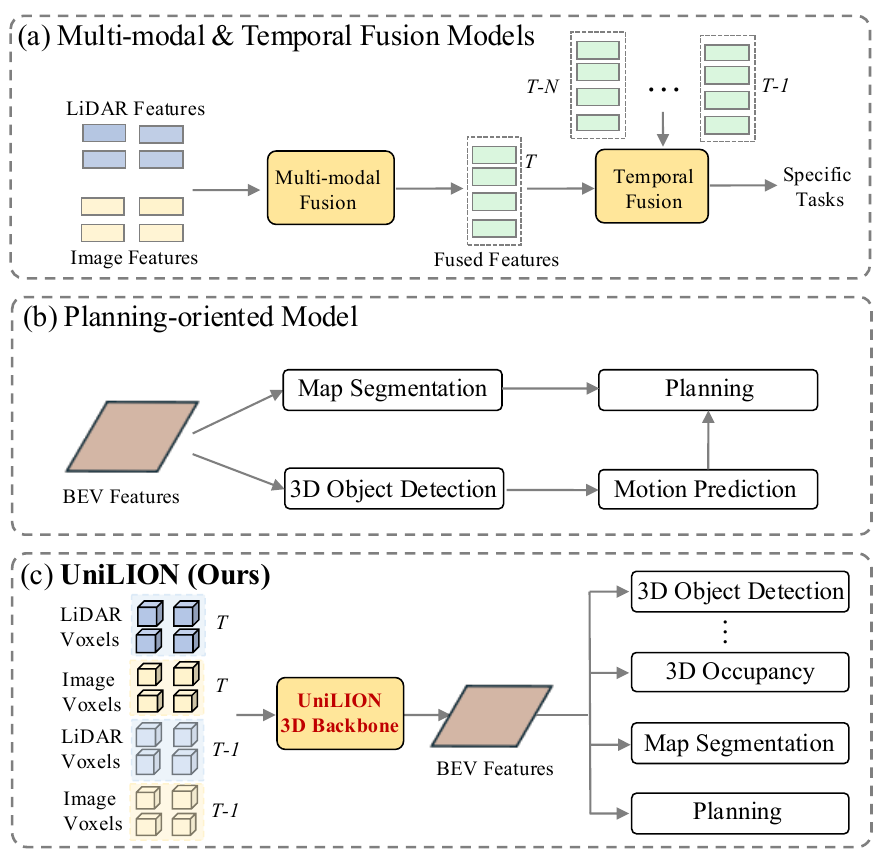}
\caption{
(a) presents the mainstream methods in implementing multi-modal fusion or temporal fusion. (b) illustrates the classic pipeline for achieving the end-to-end autonomous driving system. (c) demonstrates our method {\ourMethod}, which elegantly unifies multiple input modalities and temporal sequences into a single, versatile architecture. {\ourMethod} can seamlessly support multiple specialized
variants (\textit{i.e.}, LiDAR-only, temporal LiDAR, multi-modal, and multi-modal temporal fusion configurations) without explicit temporal or multi-modal fusion modules. Moreover, {\ourMethod} enables concurrent execution of multiple downstream tasks in a decoupled manner through a shared BEV feature representation, leveraging the comprehensive and superior feature extraction capabilities of its 3D backbone.
}
\label{fig_intro}
\vspace{-5pt}
\end{figure}

\IEEEPARstart{I}{n} autonomous driving technology, effectively processing massive heterogeneous sensor data from multi-view cameras and LiDAR sensors across both spatial and temporal dimensions, is critical for achieving robust perception, prediction, and even planning in complex driving scenarios.

As shown in Fig.~\ref{fig_intro}~(a), for spatial multi-modal fusion, classical methods~\cite{huang2020epnet,liu2022epnet++,yang2022deepinteraction} establish explicit geometric correspondences between multi-view images and LiDAR point clouds through point-wise or voxel-wise alignment to enable effective cross-modal information interaction. More recently, BEV-based fusion methods~\cite{liu2022bevfusion,liang2022bevfusion} adopt a unified spatial representation strategy by transforming heterogeneous sensor features into Bird's-Eye-View (BEV) space and then fusing multi-modal BEV representations through concatenation or attention mechanisms. For temporal modeling, recent approaches~\cite{hou2023query,koh2023mgtanet,chen2022mppnet,li2022bevformer,cai2023bevfusion4d,wang2023exploring,han2024exploring} incorporate temporal information either by geometrically aligning BEV features across frames or by employing attention-based query fusion.
Therefore, existing approaches typically require specialized multi-modal and temporal fusion modules, leading to a more complex system architecture. 

Moreover, as shown in Fig.~\ref{fig_intro}~(b), handling diverse tasks such as 3D object detection, motion prediction, and planning often necessitates intricate inter-module dependencies for better performance. For instance, UniAD~\cite{hu2023planning} and VAD~\cite{jiang2023vad} employ sequential module connections with complex task inter-dependencies, where downstream tasks rely on high-dimensional latent features from upstream modules, leading to potential error propagation and optimization challenges. Building upon UniAD, FusionAD~\cite{ye2023fusionad} further incorporates camera-LiDAR fusion through additional cross-modal attention mechanisms and feature alignment modules, yet the framework still relies on specialized fusion architectures that increase system complexity. To avoid sequential dependencies, PARA-Drive~\cite{weng2024drive} adopts a multi-task parallel architecture based on a shared BEV feature, but primarily focuses on planning-oriented optimization, leading to inferior performance on fundamental 3D perception and prediction tasks. In summary, these methods either suffer from architectural complexity with specialized fusion modules and sequential dependencies, or sacrifice performance balance across different tasks.



While transformers show promise for building unified frameworks due to their flexible attention mechanisms, their quadratic complexity becomes prohibitive when processing long sequences typical in autonomous driving scenarios, such as dense point clouds with hundreds of thousands of points, extended temporal sequences, or even multiple modalities. 
This raises a key question: \textit{Can we design a unified 3D backbone that seamlessly processes different modalities and temporal information at an acceptable computational cost across multiple tasks without any explicit fusion modules?}

Linear RNNs offer a compelling solution. One of the most important properties of linear RNNs is their linear complexity with respect to sequence length, which stands in stark contrast to the quadratic complexity of transformer-based attention mechanisms. More importantly, this computational advantage unlocks new possibilities: the linear complexity allows direct concatenation of tokens from different modalities and temporal frames into a single extended sequence, facilitating comprehensive inter-modal and temporal interactions while automatically learning complementary relationships without hand-crafted fusion designs. 

Towards this goal, we propose {\ourMethod}, a unified framework that processes multi-modal and temporal information for multiple tasks through a linear RNN-based 3D backbone, Fig.~\ref{fig_intro}~(c). This approach inherently shifts the paradigm from explicit multi-modal and temporal fusion to implicit unified representation learning like large language models~(LLMs). Specifically, heterogeneous sensor streams~(\textit{i.e.}, multi-view images, LiDAR point clouds, and temporal sequences) can be effectively unified via direct token-level~(\textit{i.e.}, regarding each voxel as a token) concatenation and collectively processed by the unified 3D backbone of {\ourMethod}, eliminating the need for specialized fusion architectures.

Benefiting from the powerful capacity of our unified 3D backbone in long-range modeling, {\ourMethod} generates a compact but comprehensive BEV features that simultaneously supports multiple downstream tasks through parallel multi-task learning. This eliminates sequential dependencies while maintaining competitive performance across perception, prediction, and planning.

Last but not least, {\ourMethod} can serve as a single versatile architecture that can seamlessly adapt to multiple specialized variants~(\textit{i.e.}, LiDAR-only, temporal LiDAR, multi-modal, and multi-modal temporal settings) without requiring explicit temporal/multi-modal fusion modules. 
Therefore, once trained with multi-modal temporal data, our {\ourMethod} can be directly deployed for LiDAR-only, temporal LiDAR, or multi-modal settings during the inference stage, enabling robust operation across diverse sensor settings and fault-tolerant deployment in safety-critical applications.





In summary, building upon our previous conference work LION~\cite{liu2024lion} for linear RNN-based 3D object detection, which introduced a 3D spatial feature descriptor to enhance local spatial information capture and a voxel generation strategy to densify foreground features, this extension makes the following new contributions:

\begin{itemize}
\item \textbf{Unified Heterogeneous Inputs:} Leveraging the superior long-range modeling capability and linear computational complexity of linear group RNNs, {\ourMethod} integrates multi-view images, LiDAR point clouds, and temporal information into a unified 3D backbone through direct token concatenation, eliminating hand-crafted fusion modules and providing a more elegant, scalable solution.
\item \textbf{Unified Model:} {\ourMethod} enables parameter sharing across different input formats. Specifically, once trained with multi-modal temporal data, the same {\ourMethod} model can be directly deployed across different sensor configurations and temporal settings~(\textit{e.g.}, LiDAR-only, temporal LiDAR, or multi-modal fusion) without retraining, demonstrating exceptional adaptability to diverse operational conditions.
\item \textbf{Unified Output Representation:} {\ourMethod} highly compresses heterogeneous multi-modal and temporal information into a compact BEV feature representation that serves as a shared feature to simultaneously support diverse autonomous driving tasks through parallel multi-task learning, eliminating sequential dependencies while maintaining competitive performance across perception, prediction, and planning tasks. 
\item \textbf{Superior Performance:} {\ourMethod} achieves competitive and state-of-the-art performance across comprehensive autonomous driving tasks, including 3D perception (\textit{e.g.}, 3D object detection, tracking, occupancy prediction, BEV map segmentation), motion prediction, and end-to-end planning, demonstrating the generalization and effectiveness of our unified approach.
\end{itemize}
\section{Related Works}


\noindent \textbf{Linear RNN.}
Recurrent Neural Networks (RNNs) were initially developed to address sequential modeling problems in Natural Language Processing (NLP), such as time series prediction and speech recognition, by effectively capturing temporal dependencies in sequential data. A key advantage of RNNs is their linear computational complexity for processing sequential features, which significantly reduces computational costs when handling long sequences compared to attention-based mechanisms.
In recent years, researchers have developed advanced time-parallelizable data-dependent RNNs (referred to as linear RNNs in this paper) to overcome the quadratic computational complexity inherent in transformer architectures~\cite{qin2023hierarchically,orvieto2023resurrecting,peng2023rwkv, peng2024eagle, sun2023retentive, de2024griffin, yang2023gated, gu2023mamba,sun2024learning,beck2024xlstm}. These modern linear RNN variants maintain the desirable linear complexity while enabling efficient parallel training, allowing them to achieve performance comparable to or even exceeding that of transformers on various tasks. Building upon these advances, numerous studies~\cite{alkin2024vision, duan2024vision,liang2024pointmamba, zhang2024voxel,liu2024lion} have explored the adaptation of linear RNN operators to diverse 2D and 3D computer vision applications. 
Particularly for large-scale outdoor 3D scenes, linear RNNs demonstrate superior capability in achieving long-range modeling with lower computational overhead compared to transformer-based approaches, resulting in enhanced performance for autonomous driving perception tasks.

\noindent \textbf{Multi-modal Temporal Fusion.}
Multi-modal fusion~\cite{liu2022epnet++,liu2022bevfusion,jiao2022msmdfusion,yang2022deepinteraction,chen2022autoalignv2,chen2023focalformer3d,yin2021multimodal} and temporal fusion~\cite{chen2022mppnet,yin2021center,hou2023query,li2022bevformer,hou2024open,wang2023exploring,han2024exploring} are critical technologies to improve the performance and robustness of autonomous driving. 
For multi-modal fusion, existing approaches can be categorized into two main paradigms. The first paradigm~\cite{liang2022bevfusion,liu2022bevfusion} transforms both point clouds and image features into a unified BEV representation, enabling multi-modal fusion through feature-level integration in the BEV space. The second paradigm~\cite{huang2020epnet,liu2022epnet++, bai2022transfusion,chen2022autoalignv2,chen2023focalformer3d} adopts a projection-based interaction mechanism, where LiDAR point-wise or voxel-wise features are projected onto multi-view images to facilitate cross-modal feature interaction.
Moreover, temporal fusion provides rich contextual information across time steps to enhance feature representation quality. Early approaches~\cite{chen2022mppnet,yin2021center} concatenate historical and current input point clouds for temporal integration. More recent methods~\cite{hou2023query,li2022bevformer} perform temporal fusion at the feature level (\textit{e.g.}, BEV features and query features). To address the computational overhead of repeatedly extracting historical features, streaming temporal fusion approaches~\cite{wang2023exploring,han2024exploring,hou2024open} have been proposed to enable efficient temporal integration over extended sequences in a streaming manner.
In contrast, our {\ourMethod} elegantly unifies multiple input modalities and temporal sequences into a single, versatile architecture. {\ourMethod} seamlessly accommodates multiple specialized configurations (\textit{i.e.}, LiDAR-only, temporal LiDAR, multi-modal, and multi-modal temporal fusion variants) without requiring explicit temporal or multi-modal fusion modules.

\noindent \textbf{3D Perception.}
3D perception is the basis of following prediction and planning tasks. From the perspective of targets, there are 3D object detection, 3D multi-object tracking, BEV map segmentation, and 3D occupancy prediction tasks. For 3D detection, point-based methods take the raw points~\cite{shi2019pointrcnn,yang2019std,qi2018frustum,qi2019deep,zhang2022not,chen2022sasa,yang20203dssd} as inputs and implement PointNets~\cite{qi2017pointnet,qi2017pointnet++} to obtain fine-grained geometric information. Voxel-based  methods~\cite{yan2018second,shi2020points,dong2022mssvt,liu2020tanet,deng2021voxel,yin2021center,guan2022m3detr,shi2020pv,shi2021pv,yang2023pvt,zhang2024safdnet,Fan_2022_CVPR,Sun2022SWFormerSW, liu2023flatformer,wang2023dsvt,wang2022cagroup3d} quantize the input irregular point clouds into regular 3D voxels for feature extraction and then convert 3D features into BEV (Bird's-eye-View) features for 3D detection. For 3D multi-object tracking, some approaches~\cite{yin2021center,zeng2021cross,ding20233dmotformer,kim2021eagermot} adopt a tracking-by-detection paradigm to track objects based on detection results, and other methods~\cite{gu2023vip3d,sun2024sparsedrive} adopt an end-to-end paradigm to jointly optimize detection and tracking. For BEV map segmentation, existing methods~\cite{li2022bevformer,liu2022bevfusion} use 2D convolution on the BEV features to predict the segmentation mask. For 3D occupancy prediction, some methods~\cite{liu2024fully,wang2024panoocc} lift multi-view image features into 3D volume features, and then adopt 3D convolution on the 3D volume features to predict the result of each voxel.


\noindent \textbf{Motion Prediction and Planning.}
Motion prediction and planning tasks involve forecasting the future trajectories of surrounding objects and the ego vehicle, respectively. Current research in autonomous driving systems can be broadly categorized into two architectural paradigms. The first paradigm employs modular architectures~\cite{hu2023planning,jiang2023vad,li2024bevformer,sun2024sparsedrive,yang2024deepinteraction++} that decompose the autonomous driving pipeline into a series of components, where planning modules explicitly depend on the motion prediction output. The second paradigm employs parallel architectures~\cite{weng2024drive} that leverage parallel processing architectures built upon Bird's Eye View~(BEV) representations, alleviating accumulative errors.


\begin{figure*}[t!]
\centering
\includegraphics[width=0.99\linewidth]{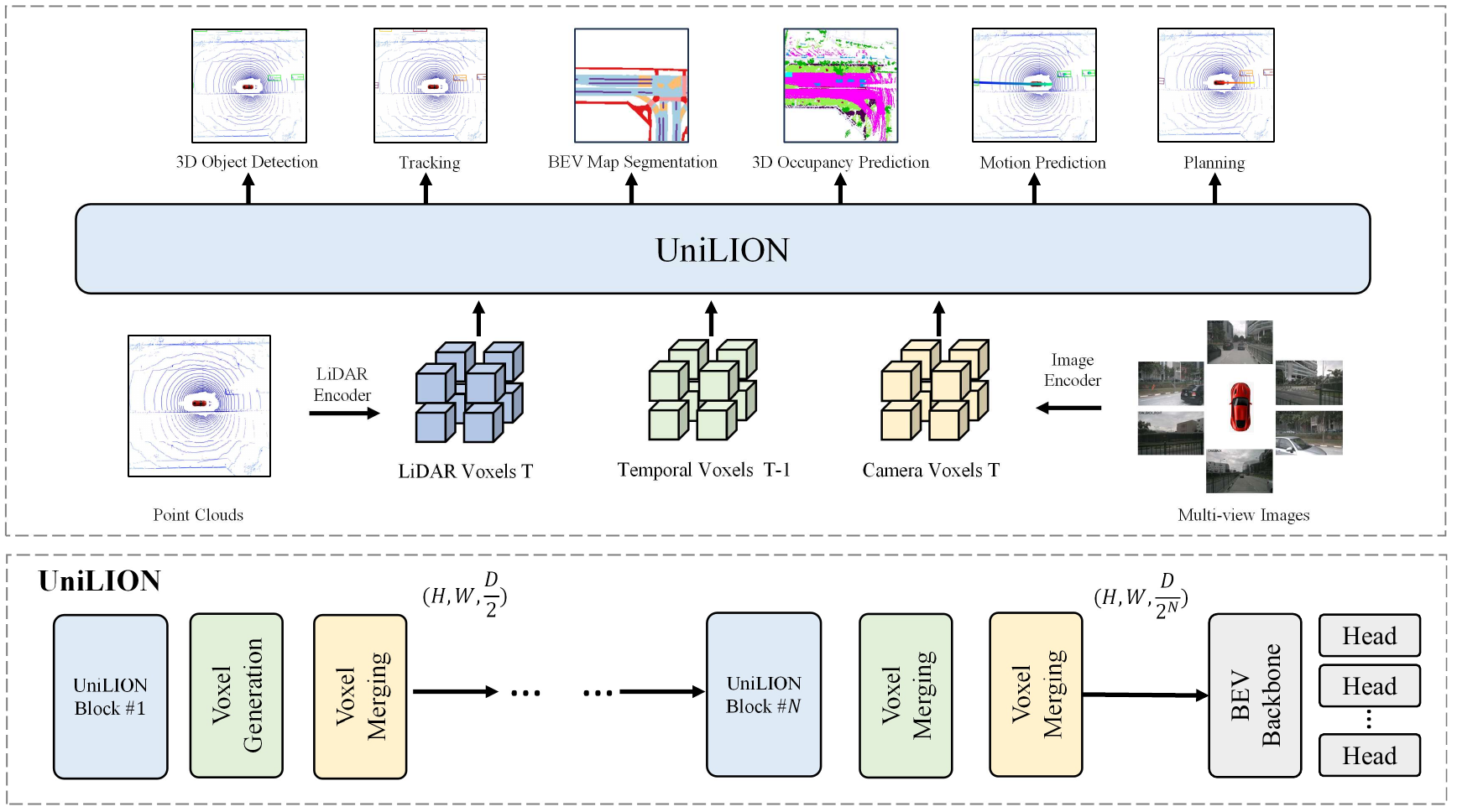}
\vspace{-5pt}
\caption{
We propose {\ourMethod}, a unified model that achieves both latent temporal fusion and multi-modal fusion in {\ourMethod} backbone by the linear group RNN, generating the unified BEV features that serve all autonomous driving tasks, including perception, prediction, and planning. 
{\ourMethod} mainly consists of $N$ {\ourMethod} blocks, each paired with a voxel generation for feature enhancement and a voxel merging for down-sampling features along the height dimension. $(H, W, D)$ indicates the shape of the 3D feature map, where $H$, $W$, and $D$ are the length, width, and height of the 3D feature map along the X-axis, Y-axis, and Z-axis. $N$ is the number of {\ourMethod} blocks. In {\ourMethod}, we first partition input multi-modal voxels into a series of equal-size groups. Then, we feed these grouped features into {\ourMethod} 3D backbone to enhance their feature representation. Finally, these enhanced features are fed into a BEV backbone to generate unified BEV features for all tasks. 
}
\label{fig_framework}
\end{figure*}

\noindent \textbf{Multi-task Learning.}
Many works have been pursuing how to process more tasks with one model (\textit{e.g.}, 3D object detection, tracking, BEV map segmentation, and 3D occupancy prediction) by multi-task learning. 
For combining the 3D detection and BEV map segmentation tasks, BEVFusion~\cite{liu2022bevfusion} adopts a separate BEV backbone to achieve the balance between the two tasks. 
When integrating 3D detection with 3D occupancy prediction, PanoOcc~\cite{wang2024panoocc} unifies feature learning and scene representation through a shared occupancy representation. 
For more complex multi-task scenarios involving 3D detection, BEV map segmentation, and 3D occupancy prediction, M3Net~\cite{chen2025m3net} introduces a task-oriented channel scaling mechanism to mitigate gradient conflicts during joint optimization.
PARA-Drive~\cite{weng2024drive} adopts a parallel multi-task architecture built upon shared BEV features for various autonomous driving tasks. However, its primary focus on planning-oriented optimization results in suboptimal performance on fundamental perception and prediction tasks.
In contrast, this paper presents {\ourMethod}, a comprehensive unified framework that simultaneously processes the complete spectrum of autonomous driving tasks (\textit{i.e.}, 3D perception, prediction, and planning). Notably, our approach achieves this unification without extra task-specific designs, relying solely on a dynamic multi-task loss mechanism (detailed in Section~\ref{method:multi_task}). Despite this simplicity, {\ourMethod} consistently delivers competitive and often state-of-the-art performance compared to specialized single-task models across all evaluated tasks.




\section{Methods}


\subsection{Overview}
In this paper, we propose a simple and effective window-based unified framework based on linear group RNN~(\textit{i.e.}, performs linear RNN for grouped features), named {\ourMethod}, which can group thousands of voxels (dozens of times more than the number of previous methods~\cite{fan2022embracing,wang2023dsvt,liu2023flatformer}) for feature interaction. 
Furthermore, {\ourMethod} can directly process temporal multi-modal voxels to achieve temporal fusion and multi-modal fusion without any additional fusion modules.
The pipeline of our {\ourMethod} is presented in Fig.~\ref{fig_framework}. {\ourMethod} consists of a LiDAR encoder, an image encoder, a unified 3D backbone, a BEV backbone, and various task heads for different tasks. 
In this paper, our key contribution lies in the unified 3D backbone for multi-modal and temporal fusion based on linear group RNN. 
In the following, we will present the details of the whole {\ourMethod}.

\subsection{LiDAR and Image Encoders}
{\ourMethod} incorporates a LiDAR encoder and an image encoder to extract LiDAR voxels and camera voxels from point clouds and multi-view images, respectively. 
For the LiDAR encoder, we convert point clouds to voxels by dynamic voxelization followed by two linear layers to generate LiDAR voxel features. 
For the image encoder, we leverage established vision image backbones~(\textit{e.g.}, ResNet-50, Swin-Tiny) to extract multi-view image features. To project these 2D image features into 3D space, we employ a lightweight depth estimation branch consisting of three 2D convolution layers to predict pixel-wise depth values. Specifically, we select the top-$K$ depth candidates (set $K=4$ by default) based on depth estimation confidence scores. These depth candidates are then combined with camera matrices to generate camera voxels in the unified 3D coordinate system. To address spatial conflicts where multiple camera voxels occupy identical 3D positions, we merge duplicate camera voxels by element-wise summation of their features at each spatial location. Finally, the extracted LiDAR voxels and camera voxels are directly concatenated along the voxel dimension and fed into the {\ourMethod} 3D backbone.


\subsection{3D Sparse Window Partition.}
{\ourMethod} implements a 3D sparse window partition to group input voxels for feature interaction. Specifically, 
we first divide input voxels into non-overlapping 3D windows with the shape of $(S_x, S_y, S_z)$, where $S_x$, $S_y$ and $S_z$ denote the length, width, and height of the window along the X-axis, Y-axis, and Z-axis. Next, we sort voxels along the X-axis for the X-axis window partition and along the Y-axis for the Y-axis window partition, respectively.
Finally, to reduce computation cost, we partition sorted voxels into groups 
of equal size $G$ rather than windows of equal shape for feature interaction.
Due to the quadratic computational complexity of transformers, previous transformer-based methods~\cite{fan2022embracing,wang2023dsvt,liu2023flatformer} only achieve feature interaction using a small group size. In contrast, we adopt a much larger group size $G$ to obtain long-range feature interaction thanks to the linear computational complexity of the linear group RNN operators.

\begin{figure*}[t!]
\centering
\includegraphics[width=0.95\linewidth]{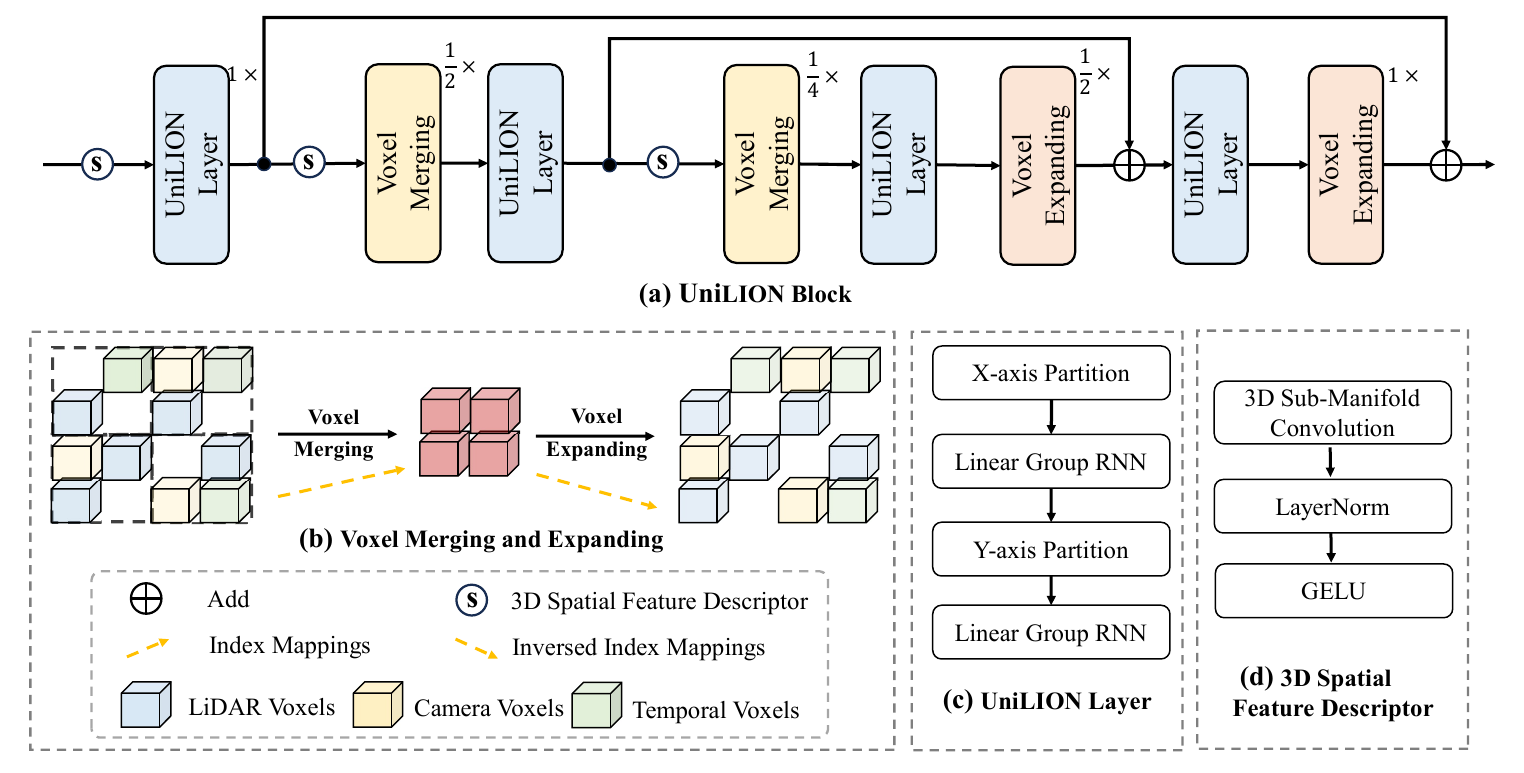}
\caption{(a) shows the structure of {\ourMethod} block, which contains four {\ourMethod} layers, two voxel merging operations, two voxel expanding operations, and three 3D spatial feature descriptors. Here, $1\times$, $\frac{1}{2}\times$, and $\frac{1}{4}\times$ indicate the resolution of 3D feature map as $(H,W,D)$, $(H/2,W/2,D/2)$ and $(H/4,W/4,D/4)$, respectively.
(b) is the illustration of voxel merging for voxel down-sampling and voxel expanding for voxel up-sampling. We use voxel merging to merge input LiDAR voxels, camera voxels, and temporal voxels to achieve multi-modal fusion and temporal fusion. (c) presents the structure of {\ourMethod} layer. (d) shows the details of the 3D spatial feature descriptor. 
}
\label{fig_block}
\vspace{-5pt}
\end{figure*}

\subsection{{\ourMethod} Block}
The {\ourMethod} block is the core component of our approach, which involves {\ourMethod} layer for long-range feature interaction, a 3D spatial feature descriptor for capturing local 3D spatial information, voxel merging for
feature down-sampling, and voxel expanding for feature up-sampling, as shown in Fig.~\ref{fig_block}~(a). Moreover, {\ourMethod} block is a hierarchical structure to better extract multi-scale features. Next, we introduce each part of {\ourMethod} block.

\noindent \textbf{Voxel Merging and Voxel Expanding.}
To enable the network to obtain multi-scale features, our {\ourMethod} adopts a hierarchical feature extraction structure. To achieve this, we need to perform feature down-sampling and up-sampling operations in highly sparse point clouds. However, it is worth mentioning that we cannot simply apply max or average pooling or up-sampling operations as in 2D images since 3D point clouds possess irregular data formats. 
Therefore, as shown in Fig.~\ref{fig_block}~(b), we adopt voxel merging for feature down-sampling and voxel expanding for feature up-sampling in highly sparse point clouds. Specifically, for voxel merging, we calculate the down-sampled index mappings to merge voxels. In voxel expanding, we up-sample the down-sampled voxels by the corresponding inverse index mappings.

\noindent \textbf{{\ourMethod} Layer.}
In {\ourMethod} block, we apply {\ourMethod} layer to model a long-range relationship among grouped features with the help of the linear group RNN operator. Specifically, as shown in Fig.~\ref{fig_block}~(c), we provide the structure of {\ourMethod} layer, which consists of two linear group RNN operators. The first one is used to perform long-range feature interaction based on the X-axis window partition, and the second one can extract long-range feature information based on the Y-axis window partition.
Through two different window partitions, {\ourMethod} layer can obtain more sufficient feature interaction, producing a more discriminative feature representation.

\begin{figure}[t!]
	\centering
	\includegraphics[width=0.55 \linewidth]{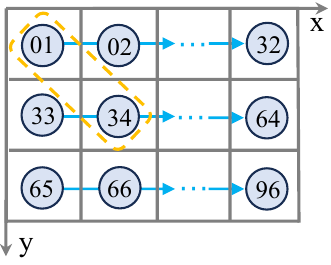}
	\caption{\small{
 The illustration of spatial information loss when flattening into 1D sequences. For example, there are two adjacent voxels in spatial position~(indexed as 01 and 34)  but are far in the 1D sequences along the X order.}
		}
        \label{fig:1d}
        \vspace{-5pt}
\end{figure}

\noindent \textbf{3D Spatial Feature Descriptor.}
Although linear RNNs have the advantages of long-range modeling with low computation cost, it is not ignorable that the spatial information might be lost when input voxel features are flattened into 1D sequential features. For example, as shown in Fig.~\ref{fig:1d}, there are two adjacent features~(\textit{i.e.}, indexed as 01 and 34) in 3D space. However, after they are flattened into 1D sequential features, the distance between them in 1D space is very far. We regard this phenomenon as a loss of 3D spatial information. 
To tackle this problem, an available manner is to increase the number of scan orders for voxel features such as VMamba~\cite{liu2024vmamba} and Vim~\cite{zhu2024vision}. However, the order of scanning is too hand-designed. Moreover, as the scanning orders increase, the corresponding computation cost also increases significantly. Therefore, it is not appropriate in large-scale sparse 3D point clouds to adopt this manner.
As shown in Fig.~\ref{fig_block}~(d), we introduce a 3D spatial feature descriptor, which consists of a 3D sub-manifold convolution, a LayerNorm layer, and a GELU activation function. Naturally, we leverage the 3D spatial feature descriptor to provide rich 3D local position-aware information for the {\ourMethod} layer. Furthermore, we place the 3D spatial feature descriptor before the voxel merging to reduce spatial information loss in the voxel merging.

\begin{figure*}[h!]
\centering
\includegraphics[width=0.99\linewidth]{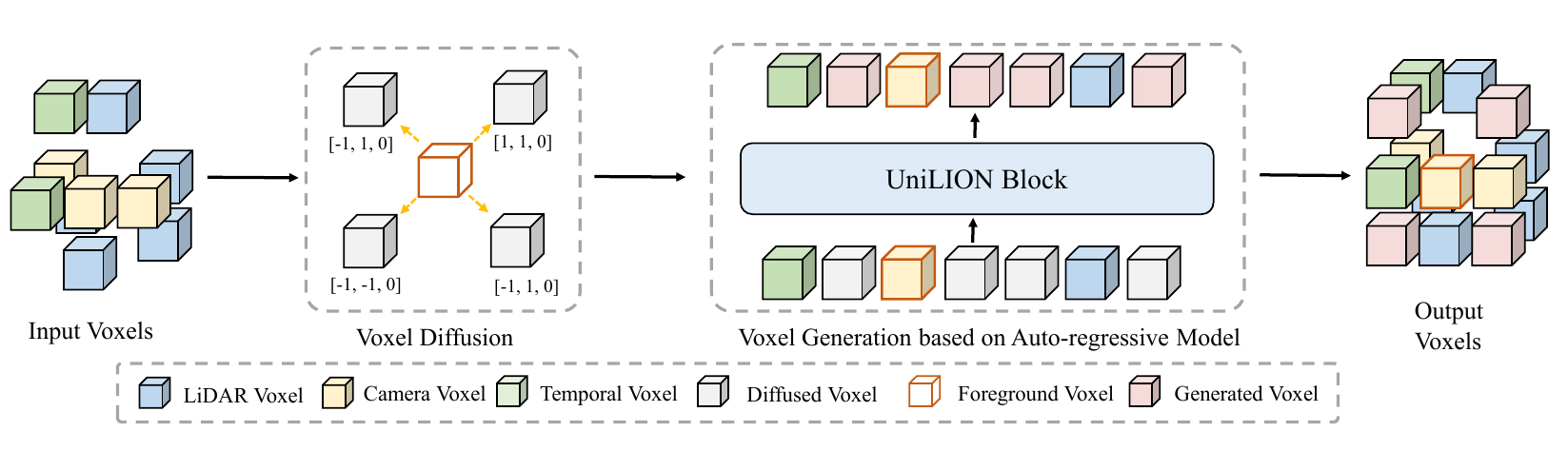}
\vspace{-5pt}
\caption{
The illustration of voxel generation. We first select the foreground voxels among LiDAR voxels, camera voxels, and temporal voxels and diffuse them along different directions. Then, we initialize the corresponding features of the diffused voxels as zeros and utilize the auto-regressive ability of the following {\ourMethod} block to generate diffused features.
}
\label{fig_vfg}
\end{figure*}

\noindent \textbf{Auto-regressive Voxel Generation.}
Although we use camera voxels and temporal voxels to compensate for LiDAR voxels in {\ourMethod} backbone, there is still the challenge of the potential information loss of implementing voxel merging in Fig.~\ref{fig_block}. Therefore, we propose a voxel generation strategy to address these issues with the help of the auto-regressive capacity of the linear group RNN. 
Specifically, for convenience, we define the corresponding coordinates of selected foreground voxel features ${F}_{m}$ as $P_m$. As shown in Fig.~\ref{fig_vfg}, 
we first obtain diffused voxels by diffusing $P_m$ with four different offsets (\textit{i.e.}, [-1,-1, 0], [1,1, 0], [1,-1, 0], and [-1,1, 0]) along the X-axis, Y-axis, and Z-axis, respectively. Then, we initialize the corresponding features of diffused voxels by all zeros. Next, we concatenate the output feature $F_i$ of the ${i}^{th}$ {\ourMethod} block with the initialized voxel features, and feed them into the subsequent $({i+1})^{th}$ {\ourMethod} block. Finally, thanks to the auto-regressive ability of the {\ourMethod} block, the diffused voxel features can be effectively generated based on other voxel features in large groups. This process is formulated as:
\begin{equation}
{F}_{p} = F_i \oplus {F}_{[-1, -1, 0]} \oplus F_{[1, 1, 0]} \oplus F_{[1, -1, 0]} \oplus F_{[-1, 1, 0]}, 
\end{equation}
\begin{equation}
F^{'}_{p} = \mathrm{Block}(F_{p}), 
\end{equation}
where $F_{[x, y, z]}$ denotes the initialized voxel features with diffused offsets of $x$, $y$, and $z$ along the X-axis, Y-axis, and Z-axis. The $\oplus$ and $\mathrm{Block}$ denote the concatenation and {\ourMethod} block, respectively.


\subsection{Unified Feature Representation}
Previous methods~\cite{liu2022bevfusion,cai2023bevfusion4d,hou2023query,koh2023mgtanet} usually adopt an additional well-designed module to achieve multi-modal or temporal fusion modules. In contrast, we aim to feed all LiDAR voxels, camera voxels, and even temporal voxels into the unified 3D backbone of {\ourMethod}, without additional multi-modal or temporal fusion modules. Thanks to the strong representation ability of {\ourMethod} 3D backbone in long-range modeling, we can adaptively model the relationship of LiDAR voxels, camera voxels, and temporal voxels.

\noindent \textbf{Multi-modal Feature Learning.} 
In autonomous driving scenarios, LiDAR point clouds and camera images exhibit strong complementarity, as LiDAR provides precise geometric structure while camera images contribute rich semantic appearance information. Therefore, {\ourMethod} aims to effectively integrate these heterogeneous modalities within a unified 3D backbone to enable mutual enhancement by leveraging the powerful long-range modeling capabilities of linear RNNs.
Specifically, given the input point clouds and multi-view images, we first quantize the point clouds into voxels and employ a voxel feature encoder (VFE) to extract these voxels, obtaining LiDAR voxels as $V_l\in \mathbb{R}^{L_l \times C}$. For multi-view images, we first employ an image backbone to extract multi-view image features. Subsequently, we adopt a depth network to predict multi-view depth maps and convert the image features into camera voxels $V_c\in \mathbb{R}^{L_c \times C}$ based on the predicted depth and the provided calibration matrix by the dataset. Here, $C$, $L_l$, and $L_c$ denote the feature channels, the number of LiDAR voxels, and the number of camera voxels, respectively. 
We then concatenate the LiDAR voxels $V_l$ and camera voxels $V_c$ to obtain multi-modal voxels $V_m \in \mathbb{R}^{(L_l + L_c) \times C}$. Remarkably, given that a spatial position in 3D space may be simultaneously occupied by both LiDAR and camera voxels, we apply the proposed voxel merging strategy to merge overlapping multi-modal voxels. Finally, we directly feed the merged multi-modal voxels $V^{'}_m \in \mathbb{R}^{L_m \times C}$ into the {\ourMethod} 3D backbone to further extract multi-modal features in 3D space, where $L_m$ represents the number of merged multi-modal voxels.


\noindent \textbf{Temporal Feature Learning.} Temporal information extraction is crucial for accurate motion prediction and trajectory planning in autonomous driving systems. Therefore, {\ourMethod} further aims to incorporate temporal information into our unified 3D backbone with the help of the powerful long-range modeling capabilities of linear RNNs.
Specifically, given the current frame multi-modal voxels $V_{T}\in \mathbb{R}^{L_{T} \times C}$, we fetch historical multi-modal voxels $V_{T-1}\in \mathbb{R}^{L_{T-1} \times C}$ from the temporal memory bank when available. To ensure spatial consistency across temporal frames, we perform spatial alignment to transform the historical voxels into the current frame's coordinate system using the transformation matrices provided by the dataset.
Subsequently, we concatenate $V_{T-1}$ and current voxels $V_{T}$ to construct temporal voxels $V_p \in \mathbb{R}^{(L_{T-1} + L_{T}) \times C}$. Here, the $L_{T-1}$ and $L_{T}$ denote the number of voxels in the $T-1$ frame and the $T$ frame, respectively. Similarly, we adopt voxel merging strategy to merge temporal voxel since multiple voxels may occupy identical 3D positions across temporal frames. Finally, we directly feed merged voxels to {\ourMethod} 3D backbone to adaptively learn temporal information. 

\subsection{Dynamic Multi-task Loss}
\label{method:multi_task}


As a unified model, {\ourMethod} consumes point clouds, multi-view images, and historical information to generate the unified BEV representation for perception, prediction, and planning of autonomous driving.
Based on the compact BEV features, we deploy task-specific heads to simultaneously output results for each task. Thanks to the modular parallel architecture, {\ourMethod} can selectively execute different tasks to reduce computational overhead during inference.
However, for multi-task training, we need to consider the problem of balancing multiple tasks. To preserve the performance of each task as much as possible, we employ a dynamic loss balancing strategy.
Specifically, given the detection loss $\mathcal{L}_{det}$, occupancy loss $\mathcal{L}_{occ}$, BEV map segmentation loss $\mathcal{L}_{map}$, motion prediction loss $\mathcal{L}_{mot}$, and planning loss $\mathcal{L}_{plan}$, we calculate the dynamic loss weight to align the loss of each task 
$\mathcal{L}_{task}$ with $\mathcal{L}_{det}$:
\begin{equation}
 w_{task} = \frac{\mathcal{L}_{det}}{\mathcal{L}_{task} + 1e^{-5}}
\end{equation}
The final loss can be formulated as:
\begin{equation}
\begin{aligned}
 \mathcal{L} =& \lambda_{1} \cdot \mathcal{L}_{det} + \lambda_{2} \cdot w_{map} \cdot \mathcal{L}_{map} + 
 \lambda_{3} \cdot w_{occ} \cdot \mathcal{L}_{occ} \\ &+ 
 \lambda_{4} \mathcal{L}_{mot} \cdot 
 + \lambda_{5} \cdot \mathcal{L}_{plan},
 \end{aligned}
\end{equation}
where $\lambda_{1}$, $\lambda_{2}$, $\lambda_{3}$, $\lambda_{4}$, and $\lambda_{5}$ are loss weights. 

\section{Experiments}

\subsection{Dataset and Evaluation Metrics} \label{exp_setup}

\noindent\textbf{Dataset.} We conduct experiments on nuScenes~\cite{caesar2020nuscenes}, a widely recognized autonomous driving benchmark featuring a perception range of up to 50 meters, with scene annotations provided at 2 Hz. The dataset comprises 1,000 scenes, partitioned into 750 training scenes, 150 validation scenes, and 150 test scenes. nuScenes provides comprehensive multi-modal data, including LiDAR-generated point clouds and images from 6 surrounding cameras with multi-view perspectives.
The dataset supports diverse autonomous driving tasks spanning perception, prediction, and planning domains, including 3D object detection, multi-object tracking, BEV map segmentation, 3D occupancy prediction, motion prediction, and trajectory planning.

\noindent \textbf{Evaluation Metrics.} We adopt task-specific metrics following established protocols: mean Average Precision (mAP) and NuScenes Detection Score (NDS) for 3D object detection; AMOTA for tracking performance; mean Intersection over Union (mIoU) for BEV map segmentation~\cite{liu2022bevfusion}; RayIoU for 3D occupancy prediction~\cite{liu2024fully}; minimum Average Displacement Error (minADE) for motion prediction~\cite{gu2023vip3d}; and L2 distance along with collision rate for planning evaluation~\cite{hu2023planning}.


\subsection{Implementation Details} \label{exp_imp_detail}

\noindent\textbf{Details of Network.} 
We evaluate four configurations of {\ourMethod} to demonstrate its versatility: LiDAR-only (L), LiDAR-Camera (LC), LiDAR-only with temporal fusion (LT), and LiDAR-Camera with temporal fusion (LCT). All variants utilize a single unified 3D backbone architecture capable of handling multiple tasks simultaneously. We set the voxel grid resolution to $(0.3\text{m}, 0.3\text{m}, 0.25\text{m})$ and employ $N = 4$ {\ourMethod} blocks with progressively refined window sizes: $(S_x, S_y, S_z) = {(13, 13, 32), (13, 13, 16), (13, 13, 8), (13, 13, 4)}$ and corresponding group sizes $G = {4096, 2048, 1024, 512}$. During voxel generation, we set the sampling ratio $r = 0.2$ to achieve an optimal balance between accuracy and computational efficiency. For camera processing, we employ two backbone configurations: Swin-Tiny~\cite{liu2021swin} as the base backbone with input resolution $384 \times 1056$, and ResNet-50~\cite{he2016deep} as the lightweight backbone with resolution $256 \times 704$. Both image backbones are pretrained on nuImages~\cite{caesar2020nuscenes}. 
For temporal modeling, we employ a streaming manner that processes 4 consecutive frames, feeding their extracted features sequentially into our {\ourMethod} backbone.

\noindent\textbf{Task-Specific Implementation Details.} We implement each task using established architectures to ensure fair evaluation. For 3D object detection, we adopt the detection head from DSVT~\cite{wang2023dsvt} and TransFusion~\cite{bai2022transfusion}. For multi-object tracking, we employ the association strategy of CenterPoint~\cite{yin2021center}. For BEV map segmentation, we utilize the BEVFusion map head~\cite{liu2022bevfusion}. For 3D occupancy prediction, we integrate the FlashOcc head~\cite{yu2023flashocc}. 
For motion prediction, we leverage detection queries processed through a single transformer decoder layer with six trajectory anchors following SparseDrive~\cite{sun2024sparsedrive}. Similarly, for trajectory planning, we employ one transformer decoder layer to predict future ego-vehicle trajectories. 
To ensure fair comparison across all tasks and prevent information leakage, we strictly avoid using any ego-status information~\cite{li2024ego} by default during training and evaluation.


\begin{table*}[t!]
\caption{Performance comparison of perception, prediction, and planning on the nuScenes validation set. $\dag$ represents using ego status information. 'L', 'C', and 'T' denote LiDAR input, camera input, and temporal information input, respectively.
}
\vspace{-5pt}
\small
\setlength{\tabcolsep}{5pt}
\resizebox{1.0\linewidth}{!}{
\centering
\begin{tabular}{l|c|c|c|c|c|c|c|c|c|c}
\toprule
\multicolumn{1}{c|}{\multirow{2}{*}{Method}} & \multirow{2}{*}{Present at} & \multirow{2}{*}{Modality} & \multicolumn{2}{c|}{\emph{Detection}} & \multicolumn{1}{c|}{\emph{Tracking}} & \multicolumn{1}{c|}{\emph{Map}} & \multicolumn{1}{c|}{\emph{Occupancy}} & \multicolumn{1}{c|}{\emph{Motion}} & \multicolumn{2}{c}{\emph{Planning}}\\
& & & NDS$\uparrow$ & mAP$\uparrow$ & AMOTA$\uparrow$ & mIoU$\uparrow$ & RayIoU$\uparrow$ & minADE (Car~/~Ped.)$\downarrow$ & L2$\downarrow$ & Col.$\downarrow$ \\
\midrule
BEVDet~\cite{huang2021bevdet} & arXiv 22 & C & 37.9 & 29.8 & \textbf{--} & \textbf{--} & \textbf{--} & \textbf{--} & \textbf{--} & \textbf{--} \\
BEVFusion-C~\cite{liu2022bevfusion} & ICRA 23 & C & 41.2 & 35.6 & \textbf{--} & 57.1 & \textbf{--} & \textbf{--} & \textbf{--} & \textbf{--} \\
VIP3D~\cite{gu2023vip3d} & CVPR 23 & CT & \textbf{--} & \textbf{--} & 21.7 & \textbf{--} & \textbf{--} & 2.05 / \textbf{--} & \textbf{--} & \textbf{--} \\
StreamPETR~\cite{wang2023exploring} & ICCV 23 & CT & 59.2 & 50.4 & \textbf{--} & \textbf{--} & \textbf{--} & \textbf{--} & \textbf{--} & \textbf{--} \\
OPEN~\cite{hou2024open} & ECCV 24 & CT & 60.6 & 51.9 & \textbf{--} & \textbf{--} & \textbf{--} & \textbf{--} & \textbf{--} & \textbf{--} \\
SparseOcc~\cite{liu2024fully} & ECCV 24 & CT & \textbf{--} & \textbf{--} & \textbf{--} & \textbf{--} & 35.1 & \textbf{--} & \textbf{--} & \textbf{--} \\
OPUS~\cite{wang2024opus} & NeurIPS 24 & CT & \textbf{--} & \textbf{--} & \textbf{--} & \textbf{--} & 41.2 & \textbf{--} & \textbf{--} & \textbf{--} \\
UniAD~\cite{hu2023planning} & CVPR 23 & CT & 49.8 & 38.0 & 35.9 & \textbf{--} & \textbf{--} & 0.71 / \textbf{--} & 0.73 & 0.61 \\
VAD~\cite{jiang2023vad} & ICCV 23 & CT & \textbf{--} & \textbf{--} & \textbf{--} & \textbf{--} & \textbf{--} & 0.72 / \textbf{--} & \textbf{--} & 0.21 \\
BEVFormer-C~\cite{li2024bevformer} & TPAMI 24 & CT & 51.7 & 41.6 & 40.5 & \textbf{--} & \textbf{--} & \textbf{--} & 0.99 & 0.70 \\
SparseDrive~\cite{sun2024sparsedrive} & ICRA 25 & CT & 58.8 & 49.6 & 50.1 & \textbf{--} & \textbf{--} & 0.60 / 0.72 & 0.61 & 0.10 \\
\midrule
CenterPoint~\cite{yin2021center} & CVPR 21 & L & 66.5 & 59.2 & \textbf{--} & \textbf{--} & \textbf{--} & \textbf{--}  & \textbf{--} & \textbf{--} \\
TransFusion-L~\cite{bai2022transfusion} & CVPR 22 & L & 70.1 & 65.5 & \textbf{--} & \textbf{--} & \textbf{--} & \textbf{--} & \textbf{--} & \textbf{--} \\
BEVFusion-L~\cite{liu2022bevfusion} & ICRA 23 & L & 69.3  &64.7  & \textbf{--} & 48.6 & \textbf{--} & \textbf{--} & \textbf{--} & \textbf{--} \\
SEED~\cite{liu2024seed} & ECCV 24 & L & 71.2 & 66.6 & \textbf{--} & \textbf{--} & \textbf{--} & \textbf{--} & \textbf{--} & \textbf{--} \\
HEDNet~\cite{zhang2023hednet} & NeurIPS 23 & L & 71.4 & 66.7 & \textbf{--} & \textbf{--} & \textbf{--} & \textbf{--} & \textbf{--} & \textbf{--} \\
LION-Mamba~\cite{liu2024lion} & NeurIPS 24 & L & 72.1 & 68.0 & \textbf{--} & \textbf{--} & \textbf{--} & \textbf{--} & \textbf{--} & \textbf{--} \\
3DMOTFormer~\cite{ding20233dmotformer} & ICCV 23 & L & \textbf{--} & \textbf{--} & 71.2 & \textbf{--} & \textbf{--} & \textbf{--} & \textbf{--} & \textbf{--}\\
\rowcolor[gray]{0.95}
{\ourMethod}~(Ours) & \textbf{--} & L & \textbf{72.3} & \textbf{67.5} & \textbf{72.6} & \textbf{71.7} & \textbf{46.8} & \textbf{--} & \textbf{--} & \textbf{--} \\
\midrule
PnPNet~\cite{liang2020pnpnet} & CVPR 20 & LT & \textbf{--} & \textbf{--} & \textbf{--} & \textbf{--} & \textbf{--} & 1.15 / \textbf{--} & \textbf{--} & \textbf{--} \\
MGTANet~\cite{koh2023mgtanet} & AAAI 23 & LT & 68.7 & 62.9 & \textbf{--} & \textbf{--} & \textbf{--} & \textbf{--} & \textbf{--} & \textbf{--} \\
QTNet~\cite{hou2023query} & NeurIPS 23 & LT & 70.9 & 66.5 & \textbf{--} & \textbf{--} & \textbf{--} & \textbf{--} & \textbf{--} & \textbf{--} \\
\rowcolor[gray]{0.95}
{\ourMethod}~(Ours) & \textbf{--} & LT & \textbf{73.0} & \textbf{68.9} & \textbf{73.3} & \textbf{72.4} & \textbf{49.6} & \textbf{0.58} / \textbf{0.39} & \textbf{0.60} & \textbf{0.27} \\
\midrule
TransFusion~\cite{bai2022transfusion} & CVPR 22 & LC & 71.3 & 67.5 & \textbf{--} & \textbf{--} & \textbf{--} & \textbf{--} & \textbf{--} & \textbf{--} \\
BEVFusion~\cite{liang2022bevfusion} & NeurIPS 22 & LC & 71.0 & 67.9 & \textbf{--} & \textbf{--} & \textbf{--} & \textbf{--} & \textbf{--} & \textbf{--} \\
DeepInteraction~\cite{yang2022deepinteraction} & NeurIPS 22 & LC & 72.6 & 69.9 & \textbf{--} & \textbf{--} & \textbf{--} & \textbf{--} & \textbf{--} & \textbf{--} \\
UniTR-Det~\cite{wang2023unitr} & ICCV 23 & LC & 73.3 & 70.5 & \textbf{--} & \textbf{--} & \textbf{--} & \textbf{--} & \textbf{--} & \textbf{--} \\
UniTR-Map~\cite{wang2023unitr} & ICCV 23 & LC & \textbf{--} & \textbf{--} & \textbf{--} & 74.7 & \textbf{--} & \textbf{--} & \textbf{--} & \textbf{--} \\
BEVFusion~\cite{liu2022bevfusion} & ICRA 23 & LC & 71.4 & 68.5 & \textbf{--} & 63.0 & \textbf{--} & \textbf{--} & \textbf{--} & \textbf{--} \\
CMT~\cite{yan2023cross} & ICCV 23 & LC & 72.9 & 70.3 & \textbf{--} & \textbf{--} & \textbf{--} & \textbf{--} & \textbf{--} & \textbf{--} \\
DAL-Large~\cite{huang2024detecting} & ECCV 24 & LC & 74.0 & 71.5 & \textbf{--} & \textbf{--} & \textbf{--} & \textbf{--} & \textbf{--} & \textbf{--} \\
EagerMOT~\cite{kim2021eagermot} & ICRA 21 & LC & \textbf{--} & \textbf{--} & 71.2 & \textbf{--} & \textbf{--} & \textbf{--} & \textbf{--} & \textbf{--} \\
AlphaTrack~\cite{zeng2021cross} & IROS 21 & LC & \textbf{--} & \textbf{--} & 73.3 & \textbf{--} & \textbf{--} & \textbf{--} & \textbf{--} & \textbf{--} \\ 
M3Net~\cite{chen2025m3net} & AAAI 25 & LC & 72.4 & 69.0 & \textbf{--} & 70.4 & \textbf{--} & \textbf{--} & \textbf{--} & \textbf{--} \\
BEVFormer-M\cite{li2024bevformer} & TPAMI 24 & LC & 73.2 & 71.2 & \textbf{--} & \textbf{--} & \textbf{--} & \textbf{--} & \textbf{--} & \textbf{--}\\
DeepInteraction++~\cite{yang2024deepinteraction++} & TPAMI 25 & LC & 73.3 & 70.6 & \textbf{--} & \textbf{--} & \textbf{--} & \textbf{--} & \textbf{--} & \textbf{--} \\
\rowcolor[gray]{0.95}
\rowcolor[gray]{0.95}
{\ourMethod}~(Ours) & \textbf{--} & LC & \textbf{74.9} & \textbf{72.2} & \textbf{76.2} & \textbf{72.3} & \textbf{50.8} & \textbf{--} & \textbf{--} & \textbf{--} \\
\midrule
QTNet~\cite{hou2023query} & NeurIPS 23 & LCT & 73.1 & 70.3 & \textbf{--} & \textbf{--} & \textbf{--} & \textbf{--} & \textbf{--} & \textbf{--}\\
BEVFusion4D~\cite{cai2023bevfusion4d} & arXiv 23 & LCT & 73.5 & 72.0 & \textbf{--} & \textbf{--} & \textbf{--} & \textbf{--} & \textbf{--} & \textbf{--}\\
FusionAD~\cite{ye2023fusionad} & arXiv 23 & LCT & 64.6 & 57.4 & 50.1 & \textbf{--} & \textbf{--} & 0.39 / \textbf{--} & 1.03 & 0.21 \\
DeepInteraction++$\dag$~\cite{yang2024deepinteraction++} & TPAMI 25 & LCT & 66.0 & 55.7 & \textbf{--} & \textbf{--} & \textbf{--} & \textbf{0.34} / \textbf{--} & 0.71 & 0.19 \\
\rowcolor[gray]{0.95}
{\ourMethod}~(Ours) & \textbf{--} & LCT & \textbf{75.4} & \textbf{73.2} & \textbf{76.5} & \textbf{73.3} & \textbf{51.3} & 0.57 / \textbf{0.37} & \textbf{0.65} & \textbf{0.18} \\
\bottomrule
\end{tabular}
 }
\label{tab:nuscenes_main_results}
\vspace{-5pt}
\end{table*}

\noindent\textbf{Details of Training Process.}
During the training stage, we employ standard data augmentation strategies including horizontal flipping, rotation, translation, scaling, and ground truth sampling~\cite{yin2021center,huang2024detecting} to enhance model robustness and mitigate overfitting.
For training {\ourMethod}, we adopt a multi-stage training strategy with carefully designed data augmentation policies. During the single-frame training stages, we use single-frame LiDAR or LiDAR-Camera data as input. First, we jointly train detection and map segmentation for 12 epochs using Class-Balanced Grouping and Sampling (CBGS)~\cite{zhu2019class} with all data augmentation techniques to ensure robust feature learning in the early training phase. Next, we train the perception model by integrating detection, map segmentation, and occupancy prediction tasks for 24 epochs with only rotation and flipping augmentations, as the ground truth occupancy annotations impose constraints on geometric transformations. After these processes, we can obtain single-frame perception models (\textit{i.e.}, LiDAR-only and LiDAR-Camera variants) of {\ourMethod}.
For temporal training, we load the weights of the corresponding single-frame pretrained perception model and feed multi-frame inputs into our model in a streaming manner for training 24 epochs to produce the temporal perception model. Subsequently,  we load and freeze the pretrained temporal perception model weights to train the motion prediction and planning tasks for 36 epochs without any data augmentation, resulting in temporal variants (\textit{i.e.}, temporal LiDAR-only and temporal LiDAR-Camera models). $\lambda_{1}$, $\lambda_{2}$, $\lambda_{3}$, $\lambda_{4}$, and $\lambda_{5}$ are set to 1, 0.5, 1, 1, and 1, respectively.


\noindent\textbf{Details of Inference Process.}
During the inference stage, unlike conventional approaches~\cite{liu2022bevfusion,cai2023bevfusion4d} that require separate models for different input modalities or temporal fusion, our single model of {\ourMethod} with multi-frame and multi-modal inputs can seamlessly support multiple specialized variants~(\textit{i.e.}, LiDAR-only, temporal LiDAR-only, LiDAR-Camera, and temporal LiDAR-Camera models), eliminating the need for separate model architectures or explicit temporal/multi-modal fusion modules. This unified paradigm maintains model elegance while still achieving high performance across different model settings. 

\begin{table*}[t!]
\caption{Detection performance on the nuScenes validation set. ‘T.L.’, ‘C.V.’, ‘Ped.’, ‘M.T.’, ‘T.C.’, and 'B.R.' are short for trailer, construction vehicle, pedestrian, motor, traffic cone, and barrier, respectively. 
}
\vspace{-5pt}
\small
\setlength{\tabcolsep}{5pt}
\resizebox{1.0\linewidth}{!}{
\centering
\begin{tabular}{l|c|c|cc|cccccccccc}
\toprule
Method & Present at & Modality & NDS$\uparrow$ & mAP$\uparrow$ & Car & Truck & Bus & T.L. & C.V. & Ped. & M.T. & Bike & T.C. & B.R.\\
\midrule
CenterPoint~\cite{yin2021center} & CVPR 21 & L & 66.5 & 59.2 & 84.9 & 57.4 & 70.7 & 38.1 & 16.9 & 85.1 & 59.0 & 42.0 & 69.8 & 68.3 \\
VoxelNeXt~\cite{chen2023voxelnext} & CVPR 23 & L & 66.7 & 60.5 & 83.9 & 55.5 & 70.5 & 38.1 & 21.1 & 84.6 & 62.8 & 50.0 & 69.4 & 69.4 \\
Uni3DETR~\cite{wang2024uni3detr}  & NeurIPS 23 & L & 68.5 & 61.7 & \textbf{--} & \textbf{--} & \textbf{--} & \textbf{--} & \textbf{--} & \textbf{--} & \textbf{--} & \textbf{--} & \textbf{--} & \textbf{--} \\
TransFusion-L~\cite{bai2022transfusion} & CVPR 22 & L & 70.1 & 65.5 & 86.9 & 60.8 & 73.1 & 43.4 & 25.2 & 87.5 & 72.9 & 57.3 & 77.2 & 70.3 \\
DSVT~\cite{wang2023dsvt} & CVPR 23 & L & 71.1 & 66.4 & 87.4 & 62.6 & 75.9 & 42.1 & 25.3 & 88.2 & 74.8 & 58.7 & 77.9 & 71.0 \\
HEDNet~\cite{zhang2023hednet} & NeurIPS 23 & L & 71.4 & 66.7 & 87.7 & 60.6 & 77.8 & 50.7 & 28.9 & 87.1 & 74.3 & 56.8 & 76.3 & 66.9 \\
LION-Mamba~\cite{liu2024lion} & NeurIPS 24 & L & 72.1 & \textbf{68.0} & 87.9 & 64.9 & 77.6 & 44.4 & 28.5 & 89.6 & 75.6 & 59.4 & 80.8 & 71.6 \\
\rowcolor[gray]{0.95}
{\ourMethod}~(Ours) & \textbf{--} & L & \textbf{72.3} & {67.5} & 88.2 & 63.8 & 78.0 & 45.7 & 28.4 & 89.1 & 75.7 & 56.0 & 79.3 & 70.3 \\
\midrule
MGTANet~\cite{koh2023mgtanet} & AAAI 23 & LT & 68.7 & 62.9 & 87.0 & 59.6 & 72.3 & 40.1 & 21.5 & 86.3 & 69.3 & 51.4 & 73.4 & 67.8 \\
QTNet~\cite{bai2022transfusion} & NeurIPS 23 & LT & 70.9 & 66.5 & 87.2 & 61.5 & 75.8 & 43.0 & 25.7 & 87.8 & 75.5 & 61.5 & 75.4 & 71.4 \\
\rowcolor[gray]{0.95}
{\ourMethod}~(Ours) & \textbf{--} & LT & \textbf{73.0} & \textbf{68.9} & 89.0 & 65.2 & 79.7 & 46.7 & 30.8 & 89.6 & 77.9 & 59.5 & 79.2 & 71.1 \\
\midrule
MVP~\cite{yin2021multimodal} & NeurIPS 21 & LC & 70.8 & 67.1 & \textbf{--} & \textbf{--} & \textbf{--} & \textbf{--} & \textbf{--} & \textbf{--} & \textbf{--} & \textbf{--} & \textbf{--} & \textbf{--} \\
TransFusion~\cite{bai2022transfusion} & CVPR 22 & LC & 71.3 & 67.5 & 87.7 & 32.2 & 75.4 & 43.7 & 27.3 & 87.7 & 75.5 & 63.5 & 77.9 & 74.2 \\
AutoAlignV2~\cite{chen2022autoalignv2} & ECCV 22 & LC & 71.2 & 67.1 & \textbf{--} & \textbf{--} & \textbf{--} & \textbf{--} & \textbf{--} & \textbf{--} & \textbf{--} & \textbf{--} & \textbf{--} & \textbf{--} \\
BEVFusion~\cite{liang2022bevfusion} & NeurIPS 22 & LC & 72.1 & 69.6 & 89.1 & 66.7 & 77.7 & 42.6 & 30.9 & 89.4 & 79.0 & 67.5 & 79.3 & 73.5 \\
DeepInteraction~\cite{yang2022deepinteraction} & NeurIPS 22 & LC & 72.6 & 69.9 & 88.5 & 64.4 & 79.2 & 44.5 & 30.1 & 88.9 & 79.0 & 67.8 & 80.0 & 76.4 \\
BEVFusion~\cite{liu2022bevfusion} & ICRA 23 & LC & 71.4 & 68.5 & \textbf{--} & \textbf{--} & \textbf{--} & \textbf{--} & \textbf{--} & \textbf{--} & \textbf{--} & \textbf{--} & \textbf{--} & \textbf{--} \\
CMT~\cite{yan2023cross} & ICCV 23 & LC & 72.9 & 70.3 & \textbf{--} & \textbf{--} & \textbf{--} & \textbf{--} & \textbf{--} & \textbf{--} & \textbf{--} & \textbf{--} & \textbf{--} & \textbf{--} \\
SparseFusion~\cite{xie2023sparsefusion} & ICCV 23 & LC & 72.8 & 70.4 & \textbf{--} & \textbf{--} & \textbf{--} & \textbf{--} & \textbf{--} & \textbf{--} & \textbf{--} & \textbf{--} & \textbf{--} & \textbf{--} \\
UniTR~\cite{wang2023unitr}  & ICCV 23 & LC & 73.3 & 70.5 & \textbf{--} & \textbf{--} & \textbf{--} & \textbf{--} & \textbf{--} & \textbf{--} & \textbf{--} & \textbf{--} & \textbf{--} & \textbf{--} \\
DAL-Large~\cite{huang2024detecting} & ECCV 24 & LC & 74.0 & 71.5 & \textbf{--} & \textbf{--} & \textbf{--} & \textbf{--} & \textbf{--} & \textbf{--} & \textbf{--} & \textbf{--} & \textbf{--} & \textbf{--} \\
M3Net~\cite{chen2025m3net} & AAAI 25 & LC & 72.4 & 69.0 & \textbf{--} & \textbf{--} & \textbf{--} & \textbf{--} & \textbf{--} & \textbf{--} & \textbf{--} & \textbf{--} & \textbf{--} & \textbf{--} \\
DeepInteraction++~\cite{yang2024deepinteraction++}  & TPAMI 25 & LC & 73.3 & 70.6 & 80.0 & 65.2 & 80.0 & 44.7 & 30.4 & 89.3 & 80.3 & 69.4 & 80.6 & 77.2 \\
\rowcolor[gray]{0.95}
{\ourMethod}~(Ours) & \textbf{--} & LC & \textbf{74.9} & \textbf{72.3} & 89.5 & 67.9 & 80.9 & 49.3 & 33.3 & 91.5 & 81.8 & 70.0 & 84.1 & 74.2 \\
\midrule
QTNet~\cite{bai2022transfusion} & NeurIPS 23 & LCT & 73.1 & 70.3 & 88.4 & 64.7 & 79.0 & 44.8 & 29.4 & 89.4 & 80.5 & 70.6 & 79.7 & 76.1 \\
BEVFusion4D~\cite{koh2023mgtanet} & arXiv 23 & LCT & 73.5 & 72.0 & 90.6 & 70.3 & 81.5 & 47.1 & 32.9 & 90.2 & 81.5 & 73.0 & 80.9 & 71.6 \\
\rowcolor[gray]{0.95}
{\ourMethod}~(Ours) & \textbf{--} & LCT & \textbf{75.4} & \textbf{73.2} & 90.3 & 69.6 & 81.7 & 49.4 & 35.6 & 91.8 & 83.5 & 71.6 & 84.8 & 73.8 \\
\bottomrule
\end{tabular}
 }
\label{tab:nuscenes_det_val_results}
\vspace{-5pt}
\end{table*}

\begin{table*}[t!]
\caption{Tracking performance on the nuScenes validation set.
}
\vspace{-5pt}
\small
\setlength{\tabcolsep}{20pt}
\resizebox{1.0\linewidth}{!}{
\centering
\begin{tabular}{l|c|c|c|ccc}
\toprule
Method & Present at & Modality & AMOTA$\uparrow$ & AMOTP$\downarrow$ & Recall$\uparrow$ & IDS$\downarrow$ \\
\midrule
VIP3D~\cite{gu2023vip3d} & CVPR 23 & CT & 21.7 & 1.625 & 0.363 & \textbf{--} \\
UniAD~\cite{hu2023planning} & CVPR 23 & CT & 35.9 & 1.320 & 0.467 & 906 \\
SparseDrive~\cite{sun2024sparsedrive} & ICRA 25 & CT & 50.1 & 1.085 & 0.601 & 632  \\
\midrule
CenterPoint~\cite{yin2021center} & CVPR 21 & L & 66.5 & 0.567 & 0.562 & 562\\
SimpleTrack~\cite{gan2023simple} & arXiv 21 & L & 69.6 & 0.547 & 0.602 & \textbf{--} \\ 
3DMOTFormer~\cite{ding20233dmotformer} & ICCV 23 & L & 71.2 & 0.515 & \textbf{--} & \textbf{341} \\
\rowcolor[gray]{0.95}
{\ourMethod}~(Ours) & \textbf{--} & L & \textbf{72.6} & \textbf{0.542} & \textbf{0.764} & 510 \\
\midrule
\rowcolor[gray]{0.95}
{\ourMethod}~(Ours) & \textbf{--} & LT & \textbf{73.3} & \textbf{0.515} & \textbf{0.765} & \textbf{537} \\
\midrule
EagerMOT~\cite{kim2021eagermot} & ICRA 21 & LC & 71.2 & 0.569 & 0.752 & 899 \\
AlphaTrack~\cite{zeng2021cross} & IROS 21 & LC & 73.3 & \textbf{--} & \textbf{--} & \textbf{--} \\
\rowcolor[gray]{0.95}
{\ourMethod}~(Ours) & \textbf{--} & LC & \textbf{76.2} & \textbf{0.499} & \textbf{0.783} & \textbf{711} \\
\midrule
\rowcolor[gray]{0.95}
{\ourMethod}~(Ours) & \textbf{--} & LCT & \textbf{76.5} & \textbf{0.477} & \textbf{0.796} & \textbf{613} \\

\bottomrule
\end{tabular}
 }
\label{tab:nuscenes_track_val_results}
\vspace{-5pt}
\end{table*}



\begin{table*}[t!]
\caption{Map segmentation performance on the nuScenes validation set.
}
\vspace{-5pt}
\small
\setlength{\tabcolsep}{10pt}
\resizebox{1.0\linewidth}{!}{
\centering
\begin{tabular}{l|c|c|c|cccccc}
\toprule
Method & Present at & Modality & mIoU$\uparrow$ & Drivable & Cross. & Walkway & Stop Line & Carpark & Divider \\
\midrule
OFT~\cite{roddick2018orthographic} & BMVC 19 & C & 42.1 & 74.0 & 35.3 & 45.9 & 27.5 & 35.9 & 33.9 \\
LSS~\cite{philion2020lift} & ECCV 20 & C & 44.4 & 75.4 & 38.8 & 46.3 & 30.3 & 39.1 & 36.5 \\
CVT~\cite{zhou2022cross} & CVPR 22 & C & 40.2 & 74.3 & 36.8 & 39.9 & 25.8 & 35.0 & 29.4 \\
M2BEV~\cite{xie2204m2bev} & arXiv 22 & C & \textbf{--} & 77.2 & \textbf{--} & \textbf{--} & \textbf{--} & \textbf{--} & 40.5 \\
BEVFusion~\cite{liu2022bevfusion} & ICRA 23 & C & 56.6 & 81.7 & 54.8 & 58.4 & 47.4 & 50.7 & 46.4 \\
\midrule
CenterPoint~\cite{yin2021center} & ECCV 21 & L & 48.6 & 75.6 & 48.4 & 57.5 & 36.5 & 31.7 & 41.9 \\
DSVT~\cite{wang2023dsvt} & CVPR 23 & L & 68.0 & 87.6 & 67.2 & 72.7 & 59.7 & 62.7 & 58.2 \\
\rowcolor[gray]{0.95}
{\ourMethod}~(Ours) & \textbf{--} & L & \textbf{71.7} & \textbf{90.0} & \textbf{73.2} & \textbf{77.0} & \textbf{64.2} & \textbf{61.1} & \textbf{64.5} \\
\midrule
\rowcolor[gray]{0.95}
{\ourMethod}~(Ours) & \textbf{--} & LT & \textbf{72.4} & \textbf{90.5} & \textbf{73.7} & \textbf{78.1} & \textbf{64.9} & \textbf{62.4} & \textbf{65.0} \\
\midrule
MVP~\cite{yin2021multimodal} & NeurIPS 21 & LC & 49.0 & 76.1 & 48.7 & 57.0 & 36.9 & 33.0 & 42.2 \\
BEVFusion~\cite{liu2022bevfusion} & ICRA 23 & LC & 62.7 & 85.5 & 60.5 & 67.6 & 52.0 & 57.0 & 53.7 \\
M3Net~\cite{chen2025m3net} & AAAI 25 & LC & 70.4 & \textbf{90.3} & 69.6 & 75.8 & 63.4 & 62.3 & 61.1 \\
\rowcolor[gray]{0.95}
{\ourMethod}~(Ours) & \textbf{--} & LC & \textbf{72.3} & 90.2 & \textbf{73.1} & \textbf{76.8} & \textbf{64.6} & \textbf{64.4} & \textbf{64.8} \\
\midrule
\rowcolor[gray]{0.95}
{\ourMethod}~(Ours) & \textbf{--} & LCT & \textbf{73.3} & \textbf{91.2} & \textbf{74.5} & \textbf{78.4} & \textbf{65.4} & \textbf{64.9} & \textbf{65.5}\\
\bottomrule
\end{tabular}
 }
\label{tab:nuscenes_map_val_results}
\end{table*}


\begin{table*}[t!]
\caption{Occupancy prediction performance on the Occ3D-nuScenes validation set.
}
\vspace{-5pt}
\small
\setlength{\tabcolsep}{16pt}
\resizebox{1.0\linewidth}{!}{
\centering
\begin{tabular}{l|c|c|c|ccc} 
\toprule
Method & Present at & Modality & RayIoU$\uparrow$ & RayIoU$_{\mathrm{1m}}$ & RayIoU$_{\mathrm{2m}}$ & RayIoU$_{\mathrm{4m}}$ \\
\midrule
RenderOcc~\cite{pan2024renderocc} & ICRA 24 & C & 19.5 & 13.4 & 19.6 & 25.5 \\
SimpleOcc~\cite{gan2023simple} & TIV 24 & C & 22.5 & 17.0 & 22.7 & 27.9 \\
BEVFormer~\cite{li2022bevformer} & ECCV 22 & CT & 32.4 & 26.1 & 32.9 & 38.0 \\
BEVDet-Occ~\cite{huang2021bevdet} & arXiv 22 & CT & 32.6 & 26.6 & 33.1 & 38.2 \\
FB-Occ~\cite{li2023fb} & arXiv 23 & CT & 33.5 & 26.7 & 34.1 & 39.7 \\
FlashOcc~\cite{yu2023flashocc} & arXiv 23 & CT & \textbf{--} & \textbf{--} & \textbf{--} & \textbf{--} \\
SparseOcc~\cite{liu2024fully} & ECCV 24 & CT & 36.1 & 30.2 & 36.8 & 41.2 \\
OPUS~\cite{wang2024opus} & NeurIPS 24 & CT & 41.2 & 34.7 & 42.1 & 46.7 \\
\midrule
\rowcolor[gray]{0.95}
{\ourMethod}~(Ours) & \textbf{--} & L & \textbf{46.7} & \textbf{43.1} & \textbf{47.3} & \textbf{49.9} \\
\rowcolor[gray]{0.95}
{\ourMethod}~(Ours) & \textbf{--} & LT & \textbf{49.6} & \textbf{46.0} & \textbf{50.2} & \textbf{52.7} \\
\rowcolor[gray]{0.95}
{\ourMethod}~(Ours) & \textbf{--} & LC & \textbf{50.8} & \textbf{47.2} & \textbf{51.3} & \textbf{53.9} \\
\rowcolor[gray]{0.95}
{\ourMethod}~(Ours) & \textbf{--} & LCT & \textbf{51.3} & \textbf{47.7} & \textbf{51.9} & \textbf{54.4} \\
\bottomrule
\end{tabular}
 }
\label{tab:nuscenes_occ_val_results}
\vspace{-5pt}
\end{table*}

\begin{table*}[t!]
\caption{Motion prediction performance on the nuScenes validation set.
}
\vspace{-5pt}
\small
\setlength{\tabcolsep}{5pt}
\resizebox{1.0\linewidth}{!}{
\centering
\begin{tabular}{l|c|c|c|c|ccc}
\toprule
Method & Present at & Modality & minADE$_{Car}$ (m) $\downarrow$ & minADE$_{Ped}$ (m) $\downarrow$ & minFDE$_{Car}$ (m)$\downarrow$ & MR$_{Car}$$\downarrow$ & EPA$_{Car}$$\uparrow$\\
\midrule
VIP3D~\cite{gu2023vip3d} & CVPR 23 & CT & 2.05 & \textbf{--} & 2.84 & 0.246 & 0.226 \\
UniAD~\cite{hu2023planning} & CVPR 23 & CT & 0.71 & \textbf{--} & 1.02 & 0.151 & 0.456 \\
SparseDrive~\cite{sun2024sparsedrive} & ICRA 25 & CT & 0.60 & 0.72 & 0.96 & 0.132 & 0.555 \\
\midrule
PnPNet~\cite{liang2020pnpnet} & CVPR 20 & LT & 1.15 & \textbf{--} & 1.95 & 0.226 & 0.222 \\
\rowcolor[gray]{0.95}
{\ourMethod}~(Ours) & \textbf{--} & LT & \textbf{0.58} & \textbf{0.39} & \textbf{1.02} & \textbf{0.166} & \textbf{0.647} \\
\midrule
FusionAD~\cite{ye2023fusionad} & arXiv 23 & LCT & 0.39 & \textbf{--} & 0.62  & 0.086 & 0.626 \\
DeepInteraction++~\cite{yang2024deepinteraction++} & TPAMI 25 & LCT & \textbf{0.34} & \textbf{--} & \textbf{0.54} & \textbf{0.047} & \textbf{--} \\
\rowcolor[gray]{0.95}
{\ourMethod}~(Ours) & \textbf{--} & LCT & 0.57 & \textbf{0.37} & 0.97 & 0.163 & \textbf{0.678} \\

\bottomrule
\end{tabular}
 }
\label{tab:nuscenes_motion_val_results}
\vspace{-5pt}
\end{table*}

\subsection{Comparisons with State-of-the-art Methods} \label{exp_compare_with_sotas}

\noindent\textbf{Overall Results.} We comprehensively evaluate four variants of {\ourMethod} across six autonomous driving tasks: 3D object detection, tracking, BEV map segmentation, 3D occupancy prediction, motion prediction, and planning on the nuScenes dataset. Note that we evaluate motion prediction and planning tasks only when temporal inputs are incorporated.
As demonstrated in Table~\ref{tab:nuscenes_main_results}, {\ourMethod} achieves leading performance across all tasks within a single unified model. For the LiDAR-only configuration, {\ourMethod} delivers 72.3\% NDS and 67.5\% mAP for 3D object detection, 72.6\% AMOTA for tracking, 71.7\% mIoU for map segmentation, and 46.8\% RayIoU for occupancy prediction on the nuScenes validation set.
When incorporating temporal fusion, performance is further enhanced across all tasks: 73.0\% NDS and 68.9\% mAP for detection, 73.3\% AMOTA for tracking, 72.4\% mIoU for map segmentation, 49.6\% RayIoU for occupancy prediction, 0.58 minADE for vehicle motion prediction, 0.39 minADE for pedestrian motion prediction, and 0.27\% collision rate for planning.
For multi-modal configurations leveraging both LiDAR and camera inputs, {\ourMethod} achieves superior performance on the four core perception tasks: 74.9\% NDS and 72.2\% mAP for detection, 76.2\% AMOTA for tracking, 72.3\% mIoU for map segmentation, and 50.8\% RayIoU for occupancy prediction.
The temporal multi-modal variant represents our most capable configuration, achieving state-of-the-art or highly competitive performance across all evaluated tasks: 75.4\% NDS and 73.2\% mAP for detection, 76.5\% AMOTA for tracking, 73.3\% mIoU for map segmentation, 51.3\% RayIoU for occupancy prediction, 0.57 minADE for vehicle motion prediction, 0.37 minADE for pedestrian motion prediction, and an exceptionally low 0.18\% collision rate for planning.
Note that we do not use ego status information in the planning task.

\noindent\textbf{Results on 3D Object Detection.}
Table~\ref{tab:nuscenes_det_val_results} presents detailed 3D object detection results for {\ourMethod} on the nuScenes validation set.
Our {\ourMethod} achieves SOTA performance across four configurations: LiDAR-only, LiDAR-temporal, multi-modal, and temporal multi-modal manners. These results verify the superiority of {\ourMethod} as a unified framework for 3D object detection, demonstrating that our approach can exceed specialized single-task models while simultaneously handling multiple autonomous driving tasks.

\noindent\textbf{Results on 3D Multi-object Tracking.}
Table~\ref{tab:nuscenes_track_val_results} presents detailed tracking results for {\ourMethod} on the nuScenes validation set. Our method consistently achieves competitive performance across all configurations: 72.6\% AMOTA for LiDAR-only manner~(1.4\% improvement over previous SOTA method 3DMOTFormer~\cite{ding20233dmotformer}), 73.3\% AMOTA for LiDAR-temporal manner, 76.2\% AMOTA for multi-modal manner~(2.9\% improvement over previous SOTA method AlphaTrack~\cite{zeng2021cross}), and 76.5\% AMOTA for temporal multi-modal manner, establishing a new SOTA result. These results demonstrate the effectiveness of {\ourMethod} for 3D multi-object tracking.


\noindent\textbf{Results on BEV Map Segmentation.}
Table~\ref{tab:nuscenes_map_val_results} presents detailed map segmentation results for {\ourMethod} on the nuScenes validation set. Our method consistently achieves competitive performance across all configurations: 71.7\% mIoU for LiDAR-only (3.7\% improvement over DSVT~\cite{wang2023dsvt}), 72.4\% mIoU for LiDAR-temporal, 72.3\% mIoU for multi-modal (1.9\% improvement over the previous SOTA multi-task method M3Net~\cite{chen2025m3net}), and 73.3\% mIoU for temporal multi-modal. These results demonstrate the effectiveness of {\ourMethod} as a unified framework for map segmentation.


\noindent\textbf{Results on Occupancy Prediction.}
Table~\ref{tab:nuscenes_occ_val_results} shows the occupancy prediction performance of {\ourMethod} on the Occ3D-nuScenes~\cite{tian2023occ3d} validation set. Our approach delivers strong results across different settings: 46.7\% RayIoU for LiDAR-only, 49.6\% RayIoU for LiDAR-temporal, 50.8\% RayIoU for multi-modal input, and 51.3\% RayIoU for temporal multi-modal input, setting a new state-of-the-art benchmark. Notably, {\ourMethod} substantially outperforms existing SOTA methods, surpassing SparseOcc~\cite{liu2024fully} and OPUS~\cite{wang2024opus} by 15.2\% and 10.1\% RayIoU respectively. These findings validate the effectiveness of {\ourMethod} for occupancy prediction.


\noindent\textbf{Results on Motion Prediction.}
Table~\ref{tab:nuscenes_motion_val_results} evaluates the motion prediction capabilities of {\ourMethod} on the nuScenes validation set. Unlike many prior works that focus solely on vehicle motion, we extend our evaluation to include pedestrian trajectories. For the LiDAR-only configuration, {\ourMethod} attains 0.58 minADE for cars and 0.39 minADE for pedestrians. In the multi-modal setting, our model achieves 0.57 minADE for cars and 0.37 minADE for pedestrians. These results highlight the generalization of {\ourMethod} as an integrated approach for motion prediction across diverse object categories.

\begin{table*}[t!]
\caption{Planning performance on the nuScenes validation set. $\dag$ represents using ego status information. 
}
\vspace{-5pt}
\small
\setlength{\tabcolsep}{14pt}
\resizebox{1.0\linewidth}{!}{
\centering
\begin{tabular}{l|c|c|cccc|cccc}
\toprule
& & & \multicolumn{4}{c|}{\emph{L2 (m)$\downarrow$}} & \multicolumn{4}{c}{\emph{Col. (\%)$\downarrow$}} \\
Method & Present at & Modality & 1s & 2s & 3s & avg. & 1s & 2s & 3s & avg. \\
\midrule
UniAD~\cite{hu2023planning} & CVPR 23 & CT & 0.45 & 0.70 & 1.04 & 0.73 & 0.62 & 0.58 & 0.63 & 0.61 \\
VAD~\cite{jiang2023vad} & ICCV 23 & CT & 0.41 & 0.70 & 1.05 & 0.72 & 0.03 & 0.19 & 0.43 & 0.21 \\
PARA-Drive~\cite{weng2024drive} & CVPR 24 & CT & 0.26 & 0.59 & 1.12 & 0.66 & 0.00 & 0.12 & 0.65 & 0.26 \\
BEVFormer-C~\cite{li2024bevformer} & TPAMI 24 & CT & \textbf{--} & \textbf{--} & \textbf{--} & 0.99 & \textbf{--} & \textbf{--} & \textbf{--} & 0.70 \\
\midrule
FF~\cite{hu2021safe} & CVPR 21 & LT & 0.55 & 1.20 & 2.54 & 1.43 & 0.06 & 0.17 & 1.07 & 0.43 \\
EO~\cite{khurana2022differentiable} & ECCV 22 & LT & 0.67 & 1.36 & 2.78 & 1.60 & 0.04 & \textbf{0.09} & 0.88 & 0.33 \\
\rowcolor[gray]{0.95}
{\ourMethod}~(Ours) & \textbf{--} & LT & \textbf{0.35} & \textbf{0.67} & \textbf{1.09} & \textbf{0.70} & \textbf{0.01} & 0.20 & \textbf{0.60} & \textbf{0.27} \\
\midrule
FusionAD~\cite{ye2023fusionad} & arXiv 23 & LCT & \textbf{--} & \textbf{--} & \textbf{--} & 1.03 & 0.25 & 0.13 & 0.25 & 0.21 \\
BEVFormer-M~\cite{li2024bevformer} & TPAMI 24 & LCT & \textbf{--} & \textbf{--} & \textbf{--} & 0.95 & \textbf{--} & \textbf{--} & \textbf{--} & 0.68 \\
DeepInteraction++$\dag$~\cite{yang2024deepinteraction++} & TPAMI 25 & LCT & 0.36 & 0.67 & 1.06 & 0.70 & 0.05 & 0.15 & \textbf{0.38} & 0.19 \\ 
\rowcolor[gray]{0.95}
{\ourMethod}~(Ours) & \textbf{--} & LCT & \textbf{0.33} & \textbf{0.62} & \textbf{0.99} & \textbf{0.65} & \textbf{0.01} & \textbf{0.12} & 0.42 & \textbf{0.18}\\
\color{gray}
{\ourMethod}$\dag$~(Ours) & \color{gray} \textbf{--} & \color{gray} LCT & \color{gray} 0.18 & \color{gray} 0.37 & \color{gray} 0.65 & \color{gray} 0.55 & \color{gray} 0.01 & \color{gray} 0.02 & \color{gray} 0.14 & \color{gray} 0.06\\
\bottomrule
\end{tabular}
 }
\label{tab:nuscenes_planning_val_results}
\vspace{-5pt}
\end{table*}


\begin{table*}[t!]
\caption{Experiments for disabling multi-modal fusion and temporal fusion during inference. We do not use any masked-modal training strategy.}
\small
\centering
\setlength{\tabcolsep}{15pt}
\resizebox{1.0\linewidth}{!}{
\begin{tabular}{c|c|c|cc|c|c|c}
\toprule
\multicolumn{1}{c|}{\multirow{2}{*}{\#}} & \multicolumn{1}{c|}{\multirow{2}{*}{Training Modality}} & \multicolumn{1}{c|}{\multirow{2}{*} {Testing Modality}} & \multicolumn{2}{c|}{\emph{Detection}} & \multicolumn{1}{c|}{\emph{Tracking}} & \multicolumn{1}{c|}{\emph{Map}} & \multicolumn{1}{c}{\emph{Occupancy}} \\
& & & NDS$\uparrow$ & mAP$\uparrow$ & AMOTA$\uparrow$ & mIoU$\uparrow$ & RayIoU$\uparrow$ \\
\midrule
\textit{a} & L & L & 72.3 & 67.5 & 72.6 & 71.7 & 46.8 \\
\textit{b} & LCT & L & 70.6  & 64.6 & 70.2 & 68.6 & 43.4 \\
\midrule
\textit{c} & LT & LT & 73.0 & 68.9 & 73.3 & 72.4 & 49.6 \\
\textit{d} & LCT & LT & 70.3 & 64.0 & 70.6 & 68.7 & 42.0 \\
\midrule
\textit{e} & LC & LC & 74.9 & 72.2 & 76.2 & 72.3 & 50.8 \\
\textit{f} & LCT & LC & 74.9 & 72.3 & 76.2 & 72.2 & 50.7 \\
\midrule
\textit{g} & LCT & LCT & 75.4 & 73.2 & 76.5 & 73.3 & 51.3 \\
\bottomrule
\end{tabular}
 }
\label{tab:ab_onemodel}
\end{table*}

\begin{table*}[t!]
\caption{Comparison of different image backbones and linear RNN operators.}
\small
\centering
\setlength{\tabcolsep}{10pt}
\resizebox{1.0\linewidth}{!}{
\begin{tabular}{c|c|c|cc|c|c|c}
\toprule
\multicolumn{1}{c|}{\multirow{2}{*}{Operators}} & \multicolumn{1}{c|}{\multirow{2}{*}{Image Backbone}} & \multicolumn{1}{c|}{\multirow{2}{*}{Resolution}} & \multicolumn{2}{c|}{\emph{Detection}} & \multicolumn{1}{c|}{\emph{Tracking}} & \multicolumn{1}{c|}{\emph{Map}} & \multicolumn{1}{c}{\emph{Occupancy}}\\
& & & NDS$\uparrow$ & mAP$\uparrow$ & AMOTA$\uparrow$ & mIoU$\uparrow$ & RayIoU$\uparrow$ \\
\midrule
{\ourMethod}-RWKV~\cite{peng2023rwkv} & ResNet50~\cite{he2016deep} & $256\times704$ & \textbf{73.7} & 69.9 & 74.0 & \textbf{71.9} & 49.7 \\
{\ourMethod}-Mamba~\cite{gu2023mamba} & ResNet50~\cite{he2016deep} & $256\times704$ & 73.6 & \textbf{70.8} & \textbf{75.0} & 71.8 & \textbf{50.2} \\
{\ourMethod}-RWKV~\cite{peng2023rwkv} & Swin-Tiny~\cite{liu2021swin} & $384\times1056$ & 74.3 & 71.4 & 75.1 & 72.0 & 49.9 \\
{\ourMethod}-Mamba~\cite{gu2023mamba} & Swin-Tiny~\cite{liu2021swin} & $384\times1056$ & \textbf{74.9} & \textbf{72.2} & \textbf{76.2} & \textbf{72.3} & \textbf{50.8} \\
\bottomrule
\end{tabular}
 }
 \vspace{-5pt}
\label{tab:ab_rnn}
\end{table*}


\noindent\textbf{Results on Planning.}
For planning task, we evaluate {\ourMethod} on the nuScenes validation set, as presented in Table~\ref{tab:nuscenes_planning_val_results}. Note that incorporating ego status information leads to information leakage~\cite{li2024ego}. Therefore, {\ourMethod} deliberately excludes ego status information to mitigate this issue. Under the LiDAR configuration, {\ourMethod} achieves 0.70$m$ L2 error and 0.27\% collision rate. With multi-modal inputs, {\ourMethod} further improves performance to 0.65$m$ L2 error and 0.18\% collision rate. Owing to the enhanced representational capabilities of {\ourMethod}, our approach substantially outperforms BEVFormer-M, reducing collision rates by a considerable margin (from 0.68\% to 0.18\%). These results confirm the effectiveness and superiority of {\ourMethod} for planning.

\noindent\textbf{One Model for All.}
{\ourMethod} is a unified architecture that supports multiple modalities and tasks through a single model framework. This design inherently enables parameter sharing across heterogeneous input formats. To validate this capability, we conduct systematic studies during inference by selectively disabling specific modalities or temporal inputs from our {\ourMethod} model trained on multi-modal temporal data~(Line \textit{g}), as presented in Table~\ref{tab:ab_onemodel}.
When temporal information alone is disabled~(Line \textit{f}), our model maintains comparable performance to the standard configuration with consistent training-testing inputs~(Line \textit{e}). Remarkably, even when both temporal and camera inputs are disabled~(Line \textit{b}), our model achieves competitive performance at 70.6 NDS, surpassing both TransFusion-L~(70.1 NDS) and BEVFusion-L~(69.3 NDS).
These findings demonstrate that a single {\ourMethod} model, once trained on multi-modal temporal data, can be seamlessly deployed across diverse sensor configurations and temporal settings~(\textit{e.g.}, LiDAR-only, temporal LiDAR, or multi-modal fusion) without requiring retraining. This capability showcases exceptional adaptability to varying operational conditions while significantly enhancing the robustness and versatility of our unified framework.



\noindent\textbf{Different Image Backbones.}
In Table~\ref{tab:ab_rnn}, we provide a lightweight version {\ourMethod}, which adopts ResNet-50~\cite{he2016deep} as the image backbone under a smaller image resolution of $256 \times 704$. Compared with the base model~(Swin-tiny~\cite{liu2021swin} as image backbone with $384 \times 1056$), the lightweight version still obtains a promising performance with 73.6\% NDS, 70.8\% mAP for 3D detection, 75.0\% AMOTA for multi-object tracking, 71.8\% mIoU for map segmentation, and 50.2\% RayIoU for 3D occupancy.


\noindent\textbf{Different Linear RNN Operators.} To verify the flexibility of our framework, we evaluate another representative linear RNN operator, RWKV~\cite{peng2023rwkv}, as shown in Table~\ref{tab:ab_rnn}. While {\ourMethod}-RWKV exhibits slightly inferior performance compared to {\ourMethod}-Mamba, it still achieves superior results across multiple autonomous driving tasks, effectively demonstrating the flexibility of our framework.


\subsection{Ablation Studies} 
\label{exp_ablation_study}
To illustrate the effectiveness of {\ourMethod}, we conduct ablation studies for {\ourMethod} on the nuScenes validation set. For quick verification, we take single-frame multi-modal {\ourMethod} with ResNet50 image backbone and $256 \times 704$ resolution as the default model and evaluate the performance on 3D perception tasks~(\textit{e.g.}, 3D object detection, tracking, map segmentation, and occupancy). 


\begin{table*}[t!]
\caption{Ablation study for each component in {\ourMethod}. We train {\ourMethod} with R50 image backbone and $256 \times 704$ image size on the nuScenes dataset.}
\small
\centering
\setlength{\tabcolsep}{10pt}
\resizebox{1.0\linewidth}{!}{
\begin{tabular}{c|c|c|cc|c|c|c}
\toprule
\multicolumn{1}{c|}{\multirow{2}{*}{\#}} & \multicolumn{1}{c|}{\multirow{2}{*}{3D Spatial Feature Descriptor}} & \multicolumn{1}{c|}{\multirow{2}{*}{Voxel Generation}} & \multicolumn{2}{c|}{\emph{Detection}} & \multicolumn{1}{c|}{\emph{Tracking}} & \multicolumn{1}{c|}{\emph{Map}} & \multicolumn{1}{c}{\emph{Occupancy}} \\
& & & NDS$\uparrow$ & mAP$\uparrow$ & AMOTA$\uparrow$ & mIoU$\uparrow$ & RayIoU$\uparrow$ \\
\midrule
\textit{a} & \textbf{--}  & \textbf{--}  &72.9  &69.0  &71.9 & 70.2 & 48.4\\
\textit{b} & $\checkmark$  & \textbf{--}  &\textbf{73.6}  &69.8 &73.8 &70.7 &49.5\\
\textit{c} & \textbf{--}  & $\checkmark$ & 73.5 &70.1 &74.6 &70.3 &48.7 \\
\textit{d} & $\checkmark$  &  $\checkmark$ & \textbf{73.6} & \textbf{70.8} & \textbf{75.0} & \textbf{71.8} & \textbf{50.2} \\
\bottomrule
\end{tabular}
 }
\label{tab:ab_com}
\end{table*}

\begin{table*}[t!]
\caption{Ablation study for the dynamic loss.}
\small
\centering
\setlength{\tabcolsep}{25pt}
\resizebox{1.0\linewidth}{!}{
\begin{tabular}{c|cc|c|c|c}
\toprule
\multicolumn{1}{c|}{\multirow{2}{*}{Dynamic Loss}} & \multicolumn{2}{c|}{\emph{Detection}} & \multicolumn{1}{c|}{\emph{Tracking}} & \multicolumn{1}{c|}{\emph{Map}} & \multicolumn{1}{c}{\emph{Occupancy}}\\
& NDS$\uparrow$ & mAP$\uparrow$ & AMOTA$\uparrow$ & mIoU$\uparrow$ & RayIoU$\uparrow$ \\
\midrule
 & 73.3 & 70.7 & 74.1 & 71.2 & \textbf{50.4} \\
$\checkmark$ & \textbf{73.6} & \textbf{70.8} & \textbf{75.0} & \textbf{71.8} & 50.2 \\
\bottomrule
\end{tabular}
 }
\label{tab:ab_dyn}
\end{table*}

\begin{table*}[t!]
\caption{Ablation study for the influence of different tasks.}
\small
\centering
\setlength{\tabcolsep}{15pt}
\resizebox{1.0\linewidth}{!}{
\begin{tabular}{c|c|c|c|cc|c|c}
\toprule
 \multicolumn{1}{c|}{\multirow{2}{*}{\#}} & \multicolumn{1}{c|}{\multirow{2}{*}{Detection}} & \multicolumn{1}{c|}{\multirow{2}{*}{Map Segmentation}} & \multicolumn{1}{c|}{\multirow{2}{*}{Occupancy}} & \multicolumn{2}{c|}{\emph{Detection}} & \multicolumn{1}{c|}{\emph{Map}} & \multicolumn{1}{c}{\emph{Occupancy}}\\
& & & & NDS$\uparrow$ & mAP$\uparrow$ & mIoU$\uparrow$ & RayIoU$\uparrow$ \\
 
\midrule
\textit{a} & $\checkmark$ & \textbf{--} & \textbf{--}  & 73.9 & 71.4 & \textbf{--} & \textbf{--} \\
\textit{b} & \textbf{--} & $\checkmark$ & \textbf{--}  & \textbf{--} & \textbf{--} & 68.3 & \textbf{--} \\
\textit{c} & \textbf{--} & \textbf{--} & $\checkmark$ & \textbf{--} & \textbf{--} & \textbf{--} & 47.5 \\
\textit{d} & $\checkmark$  & $\checkmark$  & \textbf{--} & \textbf{74.0} & \textbf{71.4} & 71.7 & \textbf{--} \\
\textit{e} & $\checkmark$  & $\checkmark$  & $\checkmark$  & 73.6 & 70.8 & \textbf{71.8} & \textbf{50.2} \\
\bottomrule
\end{tabular}
 }
\label{tab:ab_task}
\end{table*}


\noindent\textbf{Each Components of {\ourMethod}.}
We validate the individual components of {\ourMethod}, including the spatial feature descriptor and voxel generation module, as shown in Table~\ref{tab:ab_com}. When integrating the 3D spatial feature descriptor into the baseline (Line \textit{a}), this setting (Line  \textit{b}) leads to promising performance improvements of 0.7\% NDS, 0.8\% mAP, 1.9\% AMOTA, 0.5\% mIoU, and 1.1\% RayIoU. This demonstrates the effectiveness of our proposed 3D spatial feature descriptor in compensating for the limited spatial modeling capability of linear RNNs.
Furthermore, the voxel generation module (Line  \textit{c}) brings performance improvements of 0.6\% NDS, 1.1\% mAP, 2.7\% AMOTA, 0.1\% mIoU, and 0.3\% RayIoU over the baseline by enhancing foreground voxel feature representations. Finally, when all components are combined (Line  \textit{d}), {\ourMethod} achieves 73.6\% NDS, 70.8\% mAP, 75.0\% AMOTA, 71.8\% mIoU, and 50.2\% RayIoU, surpassing the baseline by 0.7\% NDS, 1.8\% mAP, 3.1\% AMOTA, 1.6\% mIoU, and 1.8\% RayIoU.

\noindent\textbf{Effectiveness of Dynamic Loss.}
In Table~\ref{tab:ab_dyn}, we present experiments to validate the effectiveness of the dynamic loss mechanism. Employing dynamic loss yields consistent improvements across most tasks: 0.3\% NDS for detection, 0.9\% AMOTA for tracking, and 0.6\% mIoU for map segmentation. However, we observe a slight degradation in 3D occupancy performance. We attribute this to the dynamic loss encouraging {\ourMethod} to prioritize overall task balance, which may come at the expense of individual task optimization, particularly for the occupancy prediction task.


\noindent\textbf{Multi-task Learning.}
In Table~\ref{tab:ab_task}, we investigate the impact of joint training on the performance across different tasks. When jointly training 3D detection and map segmentation, we observe a significant performance improvement~(71.7\% mIoU vs. 68.3\% mIoU) in the map segmentation task. When the occupancy prediction task is further incorporated, our model experiences a slight performance drop in detection but achieves a substantial improvement of 2.7\% RayIoU in occupancy prediction, as the detection task could enhance 3D occupancy estimation in general. Overall, our joint training approach achieves comparable or even superior performance compared to single-task models, demonstrating the effectiveness of the compact BEV feature representation extracted by the {\ourMethod} 3D backbone.


\noindent\textbf{Robustness of Window Size and Group Size.}
A fundamental advantage of {\ourMethod} lies in its capacity for long-range dependency modeling through the integration of linear RNNs. To evaluate the generalization and parameter sensitivity of our approach, we conduct a comprehensive robustness analysis across varying window sizes and group sizes during inference, as presented in Table~\ref{tab:ab_window}.
Specifically, we assess our {\ourMethod}~(window sizes of \{$[13, 13, 32]$, $[13, 13, 16]$, $[13, 13, 8]$, and $[13, 13, 4]$\} and group sizes of \{$4096$, $2048$, $1024$, $512$\}) across different window sizes and group sizes.
Our empirical findings demonstrate that {\ourMethod} exhibits remarkable stability and consistent performance across different window and group size configurations for various downstream tasks. This indicates that {\ourMethod}  has good extrapolation ability without strongly relying on hand-crafted priors.

\begin{table*}[t!]
\caption{Ablation study for the robustness of different window sizes and group sizes.}
\small
\centering
\setlength{\tabcolsep}{7pt}
\resizebox{1.0\linewidth}{!}{
\begin{tabular}{c|c|c|cc|c|c|c}
\toprule
\multicolumn{1}{c|}{\multirow{2}{*}{\#}} & \multicolumn{1}{c|}{\multirow{2}{*}{Window Size}} & \multicolumn{1}{c|}{\multirow{2}{*}{Group Size}} & \multicolumn{2}{c|}{\emph{Detection}} & \multicolumn{1}{c|}{\emph{Tracking}} & \multicolumn{1}{c|}{\emph{Map}} & \multicolumn{1}{c}{\emph{Occupancy}}\\
& & & NDS$\uparrow$ & mAP$\uparrow$ & AMOTA$\uparrow$ & mIoU$\uparrow$ & RayIoU$\uparrow$ \\
\midrule
\textit{a} & {$[7, 7, 32]$, $[7, 7, 16]$, $[7, 7, 8]$, $[7, 7, 4]$} &{[4096, 2048, 1024, 512]} & \textbf{73.7} & \textbf{70.8} & 74.9 & 71.7 & 50.3\\
\textit{b} & {$[13, 13, 32]$, $[13, 13, 16]$, $[13, 13, 8]$, $[13, 13, 4]$} & {[4096, 2048, 1024, 512]} & \textbf{73.7} & \textbf{70.8} & 74.9 & 71.7 & 50.0 \\
\textit{c} & {$[25, 25, 32]$, $[25, 25, 16]$, $[25, 25, 8]$, $[25, 25, 4]$} & {[4096, 2048, 1024, 512]} & 73.6 & \textbf{70.8} & 75.0 & \textbf{71.8} & \textbf{50.2} \\
\textit{d} & {$[13, 13, 32]$, $[13, 13, 16]$, $[13, 13, 8]$, $[13, 13, 4]$} & {[2048, 1024, 512, 256]} & \textbf{73.7} & \textbf{70.8} & \textbf{75.3} & 71.2 & 49.7 \\
\textit{e} & {$[13, 13, 32]$, $[13, 13, 16]$, $[13, 13, 8]$, $[13, 13, 4]$} & {[4096, 2048, 1024, 512]} & 73.6 & \textbf{70.8} & 75.0 & \textbf{71.8} & \textbf{50.2} \\
\textit{f} & {$[13, 13, 32]$, $[13, 13, 16]$, $[13, 13, 8]$, $[13, 13, 4]$} & {[8192, 4096, 2048, 1024]} & 73.6 & 70.6 & 75.1 & 71.7 & 50.0\\

\bottomrule
\end{tabular}
 }
\label{tab:ab_window}
\end{table*}


\begin{table*}[t!]
\caption{Ablation study for robustness of {\ourMethod}. The first row is LiDAR-only model without misalignment and other experiments are  {\ourMethod} with multiple modality inputs. `No' means the aligned model.}
\small
\centering
\setlength{\tabcolsep}{20pt}
\resizebox{1.0\linewidth}{!}{
\begin{tabular}{c|cc|c|c|c}
\toprule
\multicolumn{1}{c|}{\multirow{2}{*}{Levels of Misalignment }} & \multicolumn{2}{c|}{\emph{Detection}} & \multicolumn{1}{c|}{\emph{Tracking}} & \multicolumn{1}{c|}{\emph{Map}} & \multicolumn{1}{c}{\emph{Occupancy}}\\
& NDS$\uparrow$ & mAP$\uparrow$ & AMOTA$\uparrow$ & mIoU$\uparrow$ & RayIoU$\uparrow$ \\
\midrule
No~(LiDAR-only) & 72.3 & 67.5 & 72.6 & 71.7 & 46.8 \\
High~(Multi-modal) & 72.8 & 69.5 & 74.0 & 71.5 & 48.8 \\
Middle~(Multi-modal) & 73.3 & 70.2 & 74.7 & 71.6 & 49.1 \\
Low~(Multi-modal) & 73.5 & 70.6 & 74.9 & 71.7 & 50.0 \\
No~(Multi-modal) & \textbf{73.6} & \textbf{70.8} & \textbf{75.0} & \textbf{71.8} & \textbf{50.2} \\
\bottomrule
\end{tabular}
 }
\label{tab:ab_robust}
\end{table*}

\noindent\textbf{Robustness of Sensor Misalignment.}
Sensor misalignment issues may occur in most autonomous driving systems. Therefore, exploring robustness to sensor misalignment is crucial for ensuring the safety of autonomous driving systems. To verify the robustness of our {\ourMethod}, we follow FBMNet~\cite{liu2024multi} to simulate sensor misalignment between LiDAR and camera modalities. Specifically, the misalignment levels of 'Low', 'Middle', and 'High' denote that the camera extrinsic matrix is rotated by 1.5°, 3.0°, and 5.0° along the vertical direction and translated by 0.15m, 0.30m, and 0.50m, respectively. 
At the low misalignment level, {\ourMethod} maintains comparable performance across different tasks compared to the aligned model. Furthermore, our {\ourMethod} achieves promising performance with moderate degradation (0.8\% NDS, 1.3\% mAP, 1.0\% AMOTA, 0.3\% mIoU, and 1.4\% RayIoU) and demonstrates strong robustness even under high-level misalignment. 
Notably, despite camera-LiDAR misalignment, the multi-modal {\ourMethod} consistently outperforms its LiDAR-only counterpart. These experiments conclusively demonstrate the robustness of {\ourMethod} to sensor misalignment challenges.

\noindent\textbf{Top-$K$ in {\ourMethod} Backbone.}
In Table~\ref{tab:ab_topk}, we study the impact of top-K depth candidates within the {\ourMethod} 3D backbone for producing camera voxels. Overall, the experimental results show a consistent performance improvement for 3D object detection as $K$ increases, indicating that incorporating additional depth candidates enhances the model's representational capacity. However, to maintain a good trade-off between computational efficiency and model performance, we empirically determine $K=4$ as the default setting for our framework.

\begin{figure*}[t!]
\centering
\includegraphics[width=0.93\linewidth]{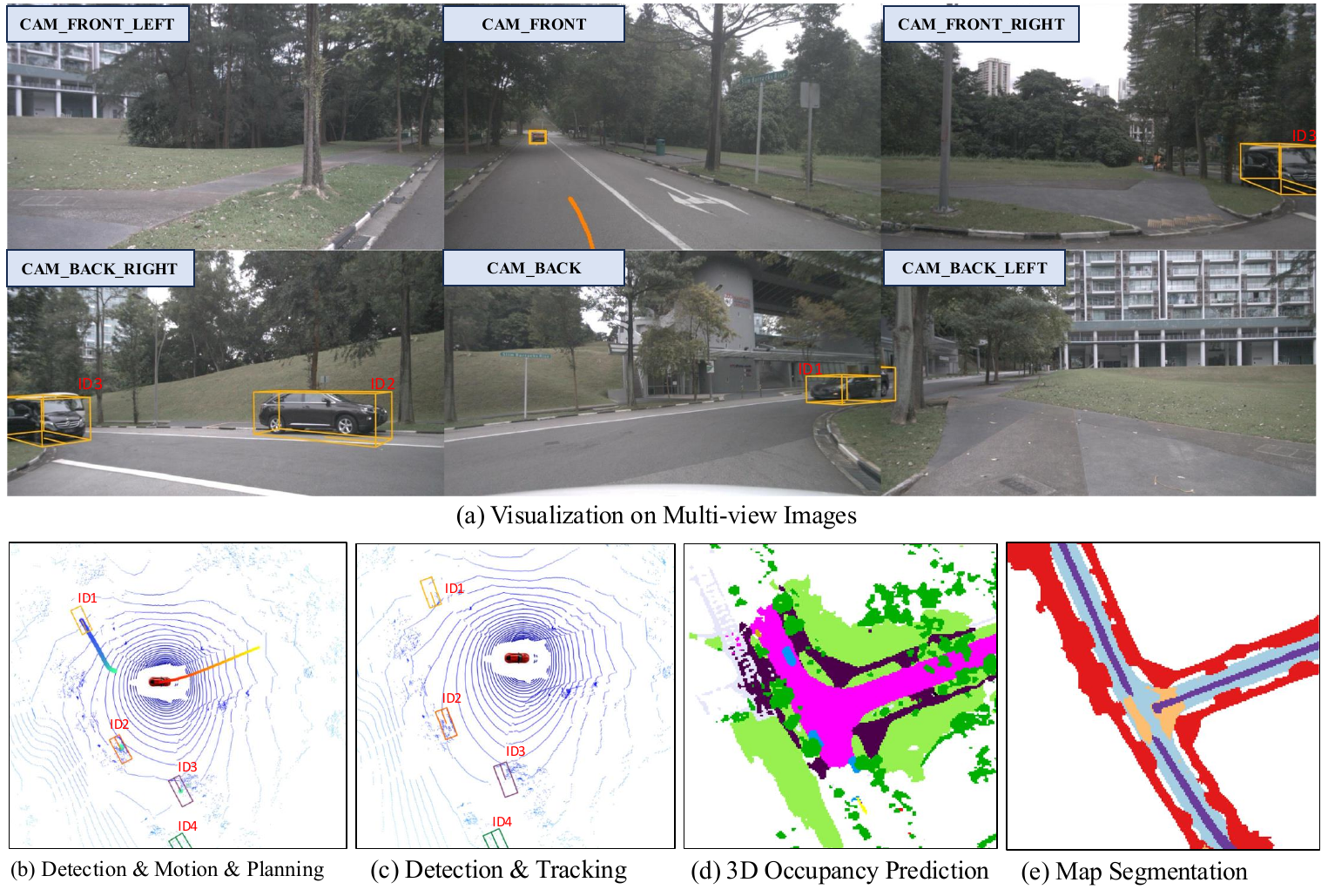}
\caption{Qualitative results of {\ourMethod} for 3D object detection, tracking, map segmentation, occupancy prediction, motion prediction, and planning on the nuScenes validation set. The first and second row is prediction results on multi-view images and on the BEV, respectively.
}
\label{fig_vis}
\end{figure*}

\begin{table*}[t!]
\caption{Ablation study for the Top-$K$ in {\ourMethod} backbone.}
\small
\centering
\setlength{\tabcolsep}{30pt}
\resizebox{1.0\linewidth}{!}{
\begin{tabular}{c|cc|c|c|c}
\toprule
\multicolumn{1}{c|}{\multirow{2}{*}{$K$}} & \multicolumn{2}{c|}{\emph{Detection}} & \multicolumn{1}{c|}{\emph{Tracking}} & \multicolumn{1}{c|}{\emph{Map}} & \multicolumn{1}{c}{\emph{Occupancy}}\\
& NDS$\uparrow$ & mAP$\uparrow$ & AMOTA$\uparrow$ & mIoU$\uparrow$ & RayIoU$\uparrow$ \\
\midrule
1 & 73.2 & 70.3 & 74.7 & 71.4 & 49.5 \\
2 & 73.5 & 70.7 & 74.8 & 71.5 & 50.2 \\
4 & 73.6 & 70.8 & 75.0 & \textbf{71.8} & \textbf{50.2} \\
8 & \textbf{73.8} & \textbf{71.1} & \textbf{75.3} & 71.7 & 50.0 \\
\bottomrule
\end{tabular}
 }
\label{tab:ab_topk}
\end{table*}

\subsection{Analysis of Visualization}
\label{exp_vis}






In this section, we present comprehensive qualitative analyses of {\ourMethod} across multiple autonomous driving tasks, including 3D object detection, tracking, map segmentation, occupancy prediction, motion prediction, and planning on the nuScenes validation set.
1)~\textbf{3D Object Detection:} As shown in Fig.~\ref{fig_vis}~(a) and~(b), we present the prediction results on the multi-view images and BEV, respectively. Thanks to a comprehensive feature representation of our {\ourMethod}, {\ourMethod} can detect a hard-to-detect object~(\textit{e.g.}, a distant object on the CAM\_FRONT view image) successfully. 
2)~\textbf{3D Object Tracking:}
As shown in Fig.~\ref{fig_vis}, we visualize the tracking results using a unique ID to distinguish different objects. {\ourMethod} can accurately associate objects cross frames in Fig.~\ref{fig_vis}(b), Fig.~\ref{fig_vis}(c).
3)~\textbf{Occupancy Prediction:}
As shown in Fig.~\ref{fig_vis}~(d), we visualize the predicted occupancy. Based on the unified BEV features, {\ourMethod} can achieve accurate 3D occupancy results. 
4)~\textbf{Map Segmentation:}
As shown in Fig.~\ref{fig_vis}~(e). {\ourMethod} can accurately segment map elements~(\textit{e.g.}, lane lines, drivable areas) and provide rich map information.
5)~\textbf{Motion and Planning:}
As shown in Fig.~\ref{fig_vis}~(b), we visualize the predictions of motion prediction and planning. For motion prediction, {\ourMethod} can accurately distinguish between moving and static objects. For planning, {\ourMethod} can produce reasonable trajectories to avoid collisions.

\section{Conclusion}
In this paper, we have presented a linear RNN-based framework {\ourMethod}, which is a unified 3D backbone that seamlessly processes different modalities and temporal information without any explicit fusion modules. Benefiting from this elegant and simple structure, this unified 3D backbone can compress different information into a compact and unified BEV representation that serves as a shared feature for seamlessly
addressing diverse autonomous driving tasks through
parallel multi-task learning. Extensive experiments verify the superiority of our {\ourMethod} in feature representation. Finally, {\ourMethod} has achieved competitive and even state-of-the-art performance across comprehensive autonomous driving tasks, including 3D perception~(3D object detection, tracking, occupancy prediction, BEV map segmentation), motion prediction, and planning, demonstrating the generalization and effectiveness of our unified approach {\ourMethod}.

{\small
	\bibliographystyle{IEEEtran}
	\bibliography{ref}
}

\end{document}